\documentclass[journal]{IEEEtran}
\ifCLASSINFOpdf
\else
\fi
\usepackage{graphicx}
\usepackage{mathptmx} 

\usepackage{latexsym}
\usepackage{times}
\usepackage{epsfig}

\usepackage{amsmath}
\usepackage{amssymb}
\usepackage{float}
\usepackage{cuted}

\usepackage[ruled,vlined]{algorithm2e}
\usepackage{multirow}
\usepackage{caption}
\usepackage[colorlinks,linkcolor=blue]{hyperref}
\newcommand{\onedot}{\ifx\@let@token.\else.\null\fi\xspace}
\newcommand{\etal}{\emph{et al}\onedot}
\newcommand{\eg}{\emph{e.g}\onedot}
\newcommand{\ie}{\emph{i.e}\onedot}

\usepackage{color}
\definecolor{yfcolor}{RGB}{255,0,0}

\begin{document}

\title{DUT: Learning Video Stabilization By Simply Watching Unstable Videos}

\author{Yufei~Xu,~\IEEEmembership{Student Member,~IEEE},
        Jing~Zhang,~\IEEEmembership{Member,~IEEE},
        Stephen~J.~Maybank,~\IEEEmembership{Fellow,~IEEE},
        and~Dacheng~Tao,~\IEEEmembership{Fellow,~IEEE}% <-this % stops a space
\thanks{This work was supported by ARC FL-170100117, DP-180103424, IH-180100002, IC-190100031. (\textit{Corresponding author: Dacheng Tao})}
\thanks{
Y. Xu, J. Zhang, and D. Tao are with the School of Computer Science, in the Faculty of Engineering, at The University of Sydney, 6 Cleveland St, Darlington, NSW 2008, Australia (email: yuxu7116@uni.sydney.edu.au; jing.zhang1@sydney.edu.au; dacheng.tao@sydney.edu.au).}% <-this % stops a space
\thanks{S. J. Maybank is with the Department of Computer Science and Information Systems, Birkbeck College, U.K (email: steve.maybank@bbk.ac.uk).}% <-this % stops a space
}

\markboth{Journal of \LaTeX\ Class Files,~Vol.~14, No.~8, August~2015}%
{Shell \MakeLowercase{\textit{et al.}}: Bare Demo of IEEEtran.cls for IEEE Journals}

\maketitle

\begin{abstract}
Previous deep learning-based video stabilizers require a large scale of paired unstable and stable videos for training, which are difficult to collect. Traditional trajectory-based stabilizers, on the other hand, divide the task into several sub-tasks and tackle them subsequently, which are fragile in textureless and occluded regions regarding the usage of hand-crafted features. In this paper, we attempt to tackle the video stabilization problem in a deep unsupervised learning manner, which borrows the divide-and-conquer idea from traditional stabilizers while leveraging the representation power of DNNs to handle the challenges in real-world scenarios.
Technically, DUT is composed of a trajectory estimation stage and a trajectory smoothing stage. In the trajectory estimation stage, we first estimate the motion of keypoints, initialize and refine the motion of grids via a novel multi-homography estimation strategy and a motion refinement network, respectively, and get the grid-based trajectories via temporal association. In the trajectory smoothing stage, we devise a novel network to predict dynamic smoothing kernels for trajectory smoothing, which can well adapt to trajectories with different dynamic patterns. We exploit the spatial and temporal coherence of keypoints and grid vertices to formulate the training objectives, resulting in an unsupervised training scheme.
Experiment results on public benchmarks show that DUT outperforms state-of-the-art methods both qualitatively and quantitatively. The source code is available at \href{https://github.com/Annbless/DUTCode}{https://github.com/Annbless/DUTCode}.
\end{abstract}

\IEEEpeerreviewmaketitle

\begin{figure*}[htbp] 
\centering 
    \includegraphics[width=\linewidth]{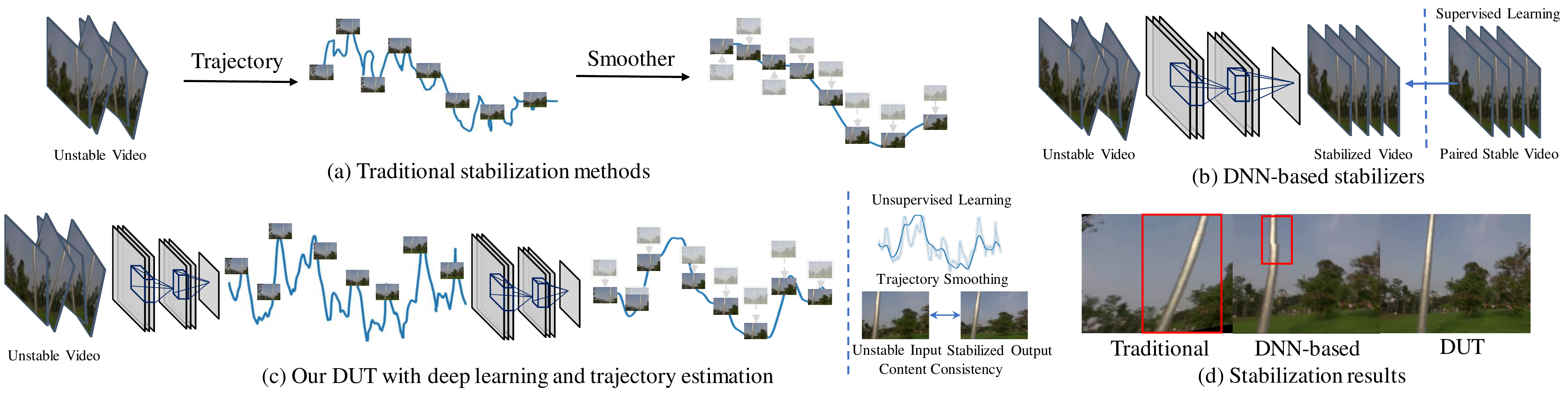}
    \caption{Comparison of our DUT and other stabilizers. (a) Traditional methods estimate the trajectory using hand-crafted features and smooth it with a fixed-kernel smoother. Stabilized frames are warped from the unstable one (marked in transparent color) according to the smoothed trajectory. (b) DNN based methods directly generate the stabilized results without explicitly modeling the smoothing process. The pipeline is hard to control and may produce distortion. (c) Our DUT stabilizes videos with explicit trajectory estimation and smoothing using DNNs. (d) Stabilized results of a traditional method \cite{liu2011subspace} (over-crop and shear), DNN based method \cite{wang2018deep} (distortion), and DUT.\label{fig:opening}}
    \label{fig:NetworkStructure}
\end{figure*}

\section{Introduction}\label{sec:introduction}

Videos captured by amateurs using hand-held cameras are usually shaky, leading to unpleasing visual experiences. Moreover, unstable videos also increase the difficulty of downstream computer vision tasks \cite{mur2015orb,bertinetto2016fully,zhang2020empowering}, \eg, object tracking and visual simultaneous localization and mapping (SLAM). Video stabilization, which aims to remove the undesirable shake and generate stabilized videos, is a fundamental image processing task and has been actively studied.

Existing video stabilizers can be roughly classified into traditional approaches~\cite{liu2009content,liu2011subspace,liu2017codingflow,zhang2017geodesic,yu2018selfie} and deep learning-based approaches~\cite{wang2018deep,zhao2020pwstablenet}, as shown in Figure~\ref{fig:opening}. Deep learning-based video stabilizers utilize deep neural networks (DNNs) to estimate the transformation from the unstable video to stable video~\cite{xu2018deep}. To effectively learn stabilization using DNNs, they usually require a large number of paired unstable and stable videos for training. Unfortunately, in real world, paired unstable and stable data are difficult to collect due to spatial and temporal synchronization~\cite{wang2018deep}. Moreover, considering that stability is a subjective visual experience that may depend on one's preference and focus, scene complexity, object moving speed, etc, there may be several reasonable stable videos that can explain a given unstable video, and vice versa. Consequently, collecting many stable videos for a given unstable video becomes more difficult. As an alternative, several stabilizers try to generate different unstable videos from one stable video using random projection matrices on each frame to mimic camera shake~\cite{yu2019robust,yu2020learning}. However, the generation of pseudo-unstable videos always ignores the influence of depth variations and dynamic objects in the stable videos. In addition, the shake patterns from random projection are different from the real-world ones. Therefore, these methods introduce a domain gap between the generated unstable videos and the real-world unstable videos. There are some other methods perform two stage stabilization by pre-stabilization first using off-the-shelf stabilizers and then post-stabilization using a new stabilizer~\cite{yu2019robust,yu2020learning}. Although the pre-stabilization process could reduce the degree of shake that may facilitate the subsequent post-stabilization process, it also introduces distortion artifacts as well as large cropping, impeding the post-stabilization process.

On the other hand, traditional video stabilization methods \cite{liu2016meshflow,liu2012video,liu2013bundled,liu2014steadyflow} stabilize videos without the requirement of paired data. Since it is difficult to directly measure the stability from the pixel values in a video, they resort to the camera trajectory as an alternative representation to account for stability and design stabilizers~\cite{liu2011subspace,liu2013bundled,liu2014steadyflow}. Specifically, these methods perform stabilization by first estimating the camera trajectory from the unstable video and then smoothing it to obtain a trajectory corresponding to an underlying stabilized video, which can be warped from the unstable one based on the transformation derived from the smoothed trajectory. These methods achieve impressive stabilization results. Nonetheless, since they rely on hand-crafted features to estimate the trajectory, the common textureless and occluded regions as well as dynamic objects in real world scenarios affect the representation ability of hand-crafted features, therefore degrading the stabilization performance~\cite{liu2009content,liu2011subspace}. Since DNNs are well known for the powerful feature representation capacity, it seems promising to use DNNs to handle the aforementioned challenging issues when estimating the trajectory. Here comes the question, \ie, can we leverage DNNs for video stabilization without using any paired data?

To answer this question, we dive into the problem formulation of the above two categories of methods. Deep learning stabilizers use paired data since they have no prior inductive bias related to stabilization and the paired data are the only source for them to learn the implicit stabilization knowledge. As for the traditional stabilizers, they tackle this task by adopting a divide-and-conquer strategy, \ie, decomposing the stabilization task into two sub-tasks of estimating the trajectory and smoothing it\footnote{The warping process to obtain the stabilized video is straightforward.}, where using the camera trajectory as a representation to account for stability indeed introduces a simple but strong prior knowledge, \ie, a smoothed trajectory implies to a stabilized video. In addition, both the estimation process and smoothing process can be performed based on the unstable video, where some other forms of prior knowledge related to stabilization is also exploited, \eg, the motion between pixels in correspondence should be small. {Based on the above analysis, we attempt to tackle the video stabilization problem in a deep unsupervised learning manner in this paper}. It borrows the divide-and-conquer idea from traditional stabilizers and leverage the representation power of DNNs to handle the challenges in real-world scenarios. 

Technically, our deep unsupervised trajectory-based stabilization method named DUT consists of a trajectory estimation stage and a trajectory smoothing stage. In the trajectory estimation stage, instead of estimating a global trajectory, we alternatively estimate the grid-based trajectories like \cite{liu2016meshflow,wang2018deep}.
Specifically, we first estimate the motion vectors of keypoints in each frame, which are more robust to the variations of noise, illumination, viewpoint, etc. The grid based motion are initialized by a novel multi-homography estimation strategy based on the keypoints and their motion. Then, we devise a novel motion refinement network to further improve the estimation accuracy.  
% Specifically, we first estimate the motion vectors of keypoints in each frame, which are more robust to the variations of noise, illumination, viewpoint, etc. Then, we devise a novel motion propagation network to propagate their motions to the grid vertices. 
In this way, we obtain the grid-based trajectories via temporal association. In the trajectory smoothing stage, we devise a novel network to predict dynamic smoothing kernels to perform trajectory smoothing, which can well adapt to different trajectories due to the diversity of scene layout, depth, and local motion. It is noteworthy that we exploit the spatial and temporal coherence of keypoints and grid vertices to formulate the training objectives, which supervise the networks to learn feature representations for each sub-tasks. In this way, our networks can be trained in an unsupervised manner. 

The main contributions of this paper are threefold:

$\bullet$ We propose the first deep unsupervised trajectory-based stabilization method (DUT) that leverages the representation power of DNNs to handle the challenges in real-world scenarios and is trained in an unsupervised manner without the requirement of paired training data.

% $\bullet$ DUT adopts an unsupervised training scheme that does not require paired unstable and stable training data.

$\bullet$ We propose a novel trajectory estimation method based on multi-homography estimation and motion refinement, which can address the issue of multiple planar motions and improve the accuracy and robustness of trajectory estimation regarding to noise, illumination change, occlusion, etc. 

$\bullet$ We propose a novel trajectory smoothing method that can predict dynamic smoothing kernels for trajectory smoothing and well adapt to trajectories with different dynamic patterns.

$\bullet$ Our DUT outperforms state-of-the-art methods both qualitatively and quantitatively on public benchmarks while being light-weight and computationally efficient.

\section{Related Work}\label{sec:related}
\begin{figure*}[htbp] 
\centering 
    \includegraphics[width=\linewidth]{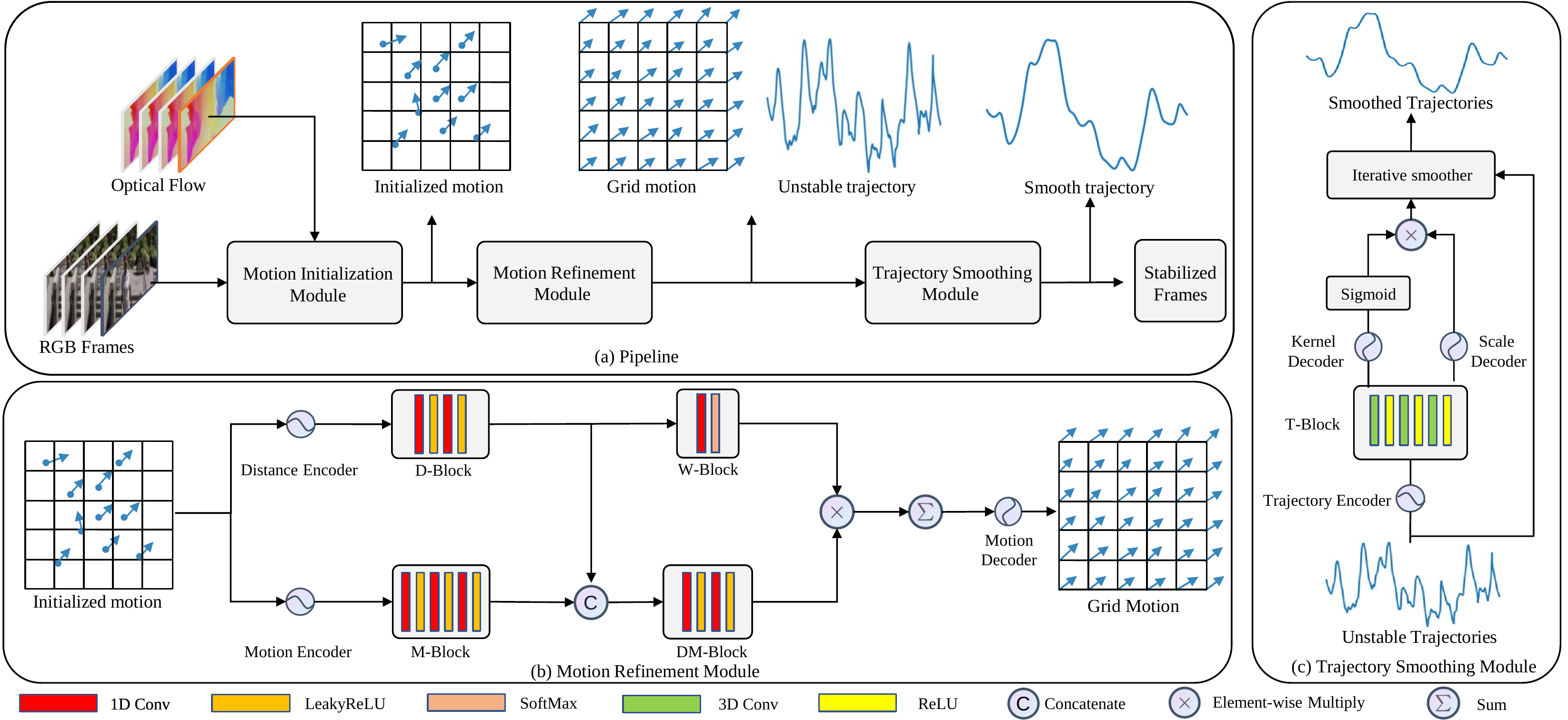}
    \caption{(a) The pipeline of our video stabilization framework, which contains three modules: keypoint detection module, motion propagation module, and trajectory smoothing module. The keypoint detection module utilizes the detector from RFNet~\cite{shen2019rf} and optical flow from PWCNet~\cite{sun2018pwc} for motion estimation. (b) The motion propagation module aims to propagate the motions from sparse keypoints to dense grid vertices. It employs two separate branches of stacked 1d convolutions to embed keypoint location and motion features, generate attention vector and attended features to predict the motion vector for each grid vertex. (c) The trajectory smoothing module aims to smooth the estimated trajectories at all grid vertices by predicting smoothing kernel weights. Three 3d convolutions are stacked to process the trajectories both spatially and temporally. The estimated kernel and scale are multiplied to form the final kernel for iterative smoothing.}
    \label{fig:NetworkStructure}
\end{figure*}

\subsection{Traditional Methods} 
Traditional methods~\cite{roberto2022survey,matsushita2006full} stabilize unstable videos by first estimating trajectories and then smoothing the trajectories for stabilization. CPW~\cite{liu2009content} utilizes structure from motion (SfM)~\cite{ullman1979interpretation} for 3D camera poses estimation, which is too fragile due to the difficulty in estimating exact camera poses and reconstruct scenes from unstable videos. \cite{wang2013spatially,gleicher2008re,grundmann2011auto} describe camera trajectories by tracking keypoints across all the frames and obtain the smoothed trajectory via optimization. {\cite{zhou2013plane} further introduces multiple-plane constraints into CPW and focuses on 3D based video stabilization. It validates the importance of recognizing different planes in video stabilization but still suffers from the fragile of 3D motion estimation.} Subspace~\cite{liu2011subspace} employs low-rank approximation for scene motion and camera motion decomposition and smooth the camera motion via polynomial path fitting. Geodesics optimization~\cite{zhang2017geodesic}, bundled path estimation~\cite{liu2013bundled}, epipolar geometry~\cite{goldstein2012video} are employed to increase robustness of the previous methods. However, these methods still need long-term tracked keypoints to estimate complete trajectories, which is challenging due to the appearance variance around keypoints. SteadyFlow~\cite{liu2014steadyflow} and MeshFlow~\cite{liu2016meshflow} indicates that the frame-to-frame motions at specific locations can be used to describe the camera motion and free the stabilizers from updating tracked or missed keypoints across the whole videos. However, they still may fail in occluded and untextured areas due to the limited expression ability of hand-crafted features. Some specific hardware~\cite{liu2012video,smith2009light,karpenko2011digital,tang2019joint,kopf2016360} is also employed to help the trajectory estimation, which imposes additional burden on the stabilization tasks. To this end, we propose to use DNN for accurate trajectory estimation and smoothing instead of relying on hand-crafted features for more robust video stabilization. With the strong expression ability of DNN models, the DUT stabilizer can better deal with unstable videos with large jitter.

\subsection{DNN based Methods}
Recently, DNN based video stabilizers have been proposed to regress unstable-to-stable transformation from data. StabNet~\cite{wang2018learning} employs spatial transform network(STN~\cite{jaderberg2015spatial}) to combines warping into the stabilization process for training using photometric loss. Similar strategy is also adopted in Xu~\etal~\cite{xu2018deep}. Although they can obtain stabilized videos, they need paired data for training and it is difficult for them to adapt to background depth variance and dynamic foreground objects, thereby producing distortion in these cases. To deal with the data shortage dilemma, some methods propose to generate unstable videos from stable videos by applying random generated projection matrices, \eg, Choi \etal proposes a neighboring frame interpolation method \cite{Choi_TOG20} based on generated unstable frames, which can generate stable videos but may introduce ghost artifacts around dynamic objects and buildings with large parallax. Such phenomena is caused by the ignorance of depth variance and dynamic objects during the generation of unstable videos. To mitigate the gaps between generated and real-life unstable videos, \cite{yu2019robust,yu2020learning} utilizes a two-stage stabilization pipeline by first utilizing traditional methods to pre stabilize the unstable videos and use DNN for further stabilization. Such design makes the performance of stabilizer is constrained both to the traditional methods' and the DNN models' performance especially in huge jitter cases.
In contrast, our DUT method learns to stabilize unstable videos from the unstable videos themselves by progressively learning to estimate unstable trajectories and then smooth the trajectories. It directly and only uses the unstable videos for training, which alleviates the data shortage problem as the unstable videos themselves are easy to collect. Besides, due to there is no domain gap between the training data and real life unstable videos, DUT can generalize to a set of videos with good performance.

\section{Methodology}\label{sec:methods}

As shown in Figure~\ref{fig:NetworkStructure}, given an unstable video as input, the DUT aims to generate a stable video by dividing this task into trajectory estimation and smoothing according to a divide-and-conquer strategy. It first estimates the grid-based trajectory by associating the motion of grid vertices, which are initialized with the motion of keypoints. The initialized motions are then further finetuned to more accurately describe the camera motion. Finally a deep learning based smoother stage is employed for a fast and stable trajectory smoothing by estimating dynamic smoothing kernels.

\subsection{Motion Initialization}
The motion initialization module (MI) aims to mitigate the effect of dynamic objects on the estimated camera motion, based on the motion of key points. Since the motion caused by camera shake is spatially coherent and the motion caused by dynamic objects is always different from this motion, we propose to enhance this coherence by using grid-based motion instead of pixel-based motion, which implicitly suppresses the effect of dynamic objects.

Denote the input video as $\left\{f_i | \forall i \in [1,E] \right\}$, where $f_i$ is the $i$th frame of the video and $E$ is the number of frames, the motion initialization stage first extracts optical flow between adjacent frames using PWCNet~\cite{sun2018pwc}, \ie,
\begin{equation}
    OF_{i} = PWC(f_i, f_{i+1}).
\end{equation}
PWCNet is a common choice in previous stabilizers~\cite{Choi_TOG20} and we use PWCNet here for a fair performance comparison. The dense pixel-based motion contains noise from dynamic objects, occlusion, etc. Previous stabilizers~\cite{yu2020learning, liu2014steadyflow} alleviates such noise using heuristic rules, \eg, predefined thresholds to exclude the influence of dynamic objects' motion. Users need to carefully choose an appropriate threshold for each unstable video, which is not satisfactory for a stabilizer than can generalize to any unstable videos. Instead of using heuristic rules, we exploit the spatial coherence of motion caused by shake to alleviate the influence of dynamic objects by estimating grid-based motion from the pixel-based motion, using sparse keypoints as a middle presentation. We adopt the RFNet~\cite{shen2019rf} as our keypoint detector to find the distinctive regions where the extracted optical flow is more reliable, \ie, 
\begin{equation}
    \{p_{ij} = RFNet\left( f_i \right) | \forall i \in [1,E], \forall j \in [1,L] \},
\end{equation}
where $p_{ij}$ is the $j$th detected keypoint on frame $f_i$, $L$ is the number of detected keypoints, which is set to 512 in this paper. Compared with the traditional keypoint detectors \cite{bay2006surf,ng2003sift,viswanathan2009features} and other deep learning based ones \cite{detone2018superpoint,ono2018lf}, RFNet can efficiently and robustly produce high-resolution response maps for selecting robust keypoints since it adopts a multi-scale and shallow network structure. For simplicity, we use the same symbol $p_{ij}$ to denote the keypoint location without causing ambiguity and the motion vectors of these keypoints can be expressed as $m_{ij}=OF_i(p_{ij})$.

\begin{figure}[htbp]
    \centering
    \includegraphics[width=\linewidth]{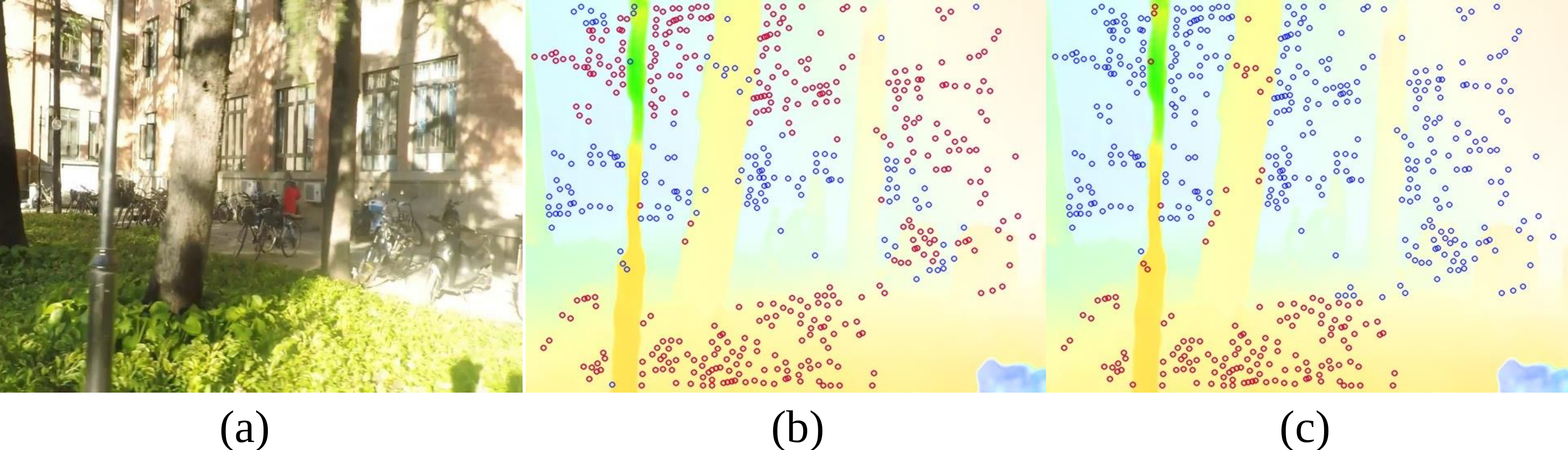}
    \caption{(a) An input image. (b) Inliers (red points) and outliers (blue points) identified using the single-homography estimation with RANSAC. (c) Two clusters of keypoints identified by our multi-homography estimation method based on K-means. The background of (b) and (c) is the visualized optical flow field of (a).}
    \label{fig:MotionKeypoint}
\end{figure}

After that, a simple yet efficient multi-homography estimation strategy is employed to initialize the grid-based motion using the keypoint-based motion. We uniformly divide each frame into a fixed set of grids, \ie, $M \times N$ grids in this paper, and use K-means to broadly segment each frame into two planes based on the keypoints motion, \ie, $C_i^c = \{\left(p_{ij}, m_{ij} \right) | \forall j \in \Lambda_i^c\}$, where $c=0$ or $1$ denotes the cluster index and $\Lambda_i^c$ is the keypoint index set belonging to the cluster $C_i^c$. Note that when the number of keypoints in any cluster is less than a threshold, \ie, 20\%$\times L=102$ in this paper, we merge it into the master cluster. Then each grid is assigned to one plane according to the majority cluster index of its neighboring keypints in $\Omega_{ik}$, which is defined as $\Omega_{ik}=\{j | \left \| d_{ijk} \right \|_2 \leq R\}$. Here, $R$ is a predefined radius, \ie, 200 pixels in this paper. Plane-aware homography are estimated based on the segmented keypoints to initialize the grid based motion, \ie, $H_i^c = Homo\left(\{p_{ij}^c|\forall j\}, \{m_{ij}^c|\forall j\}\right)$ and $\widehat{n}_{ik}=H_i^c(v_{ik}) - v_{ik}$, where $v_{ik}$ is the $k$th grid vertex on the $i$th frame and $\widehat{n}_{ik}$ is the motion of the grid vertex estimated by the initialization stage. It is worth noting that although we only use two homography estimation for motion initialization, our method can also handle frames containing more than two planes, as shown in Figure~\ref{fig:Quality}.

\subsection{Motion Refinement}

While the initialized motion can mitigate the effects of dynamic objects, the estimated motion is far from satisfactory as it does not estimate the camera motion accurately enough. To this end, we propose a CNN-based motion refinement stage to improve the accuracy of the estimated motion. Specifically, we first reconstruct the motion on detected keypoints based on the initialized gird vertex motion and calculates the residual between the reconstructed motion and the original motion estimated by PWCNet, \ie, 
\begin{equation}
    \Delta{m_{ij}} = p_{ij} + m_{ij} - H_i^c  \left( p_{ij}\right), \forall j \in \Lambda_i^c.
    \label{eq:residualmotion}
\end{equation}

The estimated motion should be accurate enough to make the residual $\Delta{m_{ij}}$ small enough for each keypoints. However, as the keypoints still may cover dynamic objects, forcing the estimated motion to minimize the residual for each keypoint may introduce extra noise. Here we explore the motion coherence property again to distinguish the keypoints belong to static backgrounds or dynamic objects based on the assumption that the motion on static background should be consistent while on dynamic objects should not obey the global coherence. Such property is utilized to train the refinement model to find a better solution for accurate motion estimation. 

Technically, we treat the sparse keypoints and dense grid vertices as two point clouds with motion and 2D location distributes and utilize several 1D convolutions to associate the keypoints' residual motion to each grid vertices to finetune the vertices' motion. Specifically, given $\Delta{m_{ij}}$ and the distance vector $d_{ijk}$ between each grid vertex $v_{ik}$ and keypoint $p_{ij}$, \ie, $d_{ijk} = p_{ij} - v_{ik}$, $k=1,\ldots,MN$, the motion refinement stage predicts the residual motion vector $\Delta{n_{ik}}$ for each vertex $v_{ik}$, \ie,
\begin{equation}
    \Delta{n_{ik}} = MR\left( \{ \left( \Delta{m_{ij}}, d_{ijk} \right) | \forall j \in [1,L] \} \right),
    \label{eq:residualmotiongrid}
\end{equation}
where $MR\left( \cdot \right)$ denotes the proposed motion refinement stage. The distance vectors and motion vectors of size $[MN,L,2]$ (after tiling) are fed into two separate encoders as shown in Figure~\ref{fig:NetworkStructure}(b). The encoded features are further embedded by several 1D convolutions. Then, the attention vector of size $[MN,L,1]$ is obtained based on the distance embeddings and used to aggregate the features from concatenated motion embeddings and distance embeddings, generating the attended feature of size $[MN,1,2]$, which is further used to predict the final dense residual motions $\Delta{n_{ik}}$ through a motion decoder. Then we can calculate the target motion vector of each vertex $v_{ik}$ as follows by referring to Eq.~\eqref{eq:residualmotion},
\begin{equation}
    {n}_{ik} = \Delta{n_{ik}} + \widehat{n}_{ik},
    \label{eq:motiongrid}
\end{equation}
The refined motion are then collected to formulate the trajectories on each grid and served as the input for the trajectory smoothing stage.

\subsection{Trajectory Smoothing (TS)}

Associating the motion vectors of corresponding grid vertices temporally formulates the trajectories that describe the camera motions, \ie, $T_k=\{ t_{ik} = \sum_{m=1}^{i}n_{mk} | \forall i \in [1,E] \}$. The key issue in the trajectory smoothing part is how to properly define the smoothness of the trajectory with respect to the different scene layouts, depths, etc. in each video. Here, we jointly use temporal and spatial coherence to characterize the properties of an ideal smooth trajectory based on grid vertices. First, we use temporal local smoothness to describe an aspect of the desired smoothness trajectory by constraining the motion between each temporal local region to be small, \ie, $L_s = \sum_{j\in\Omega_i} w_{ij}({\| \widehat{T_{ik}} - \widehat{T_{jk}} \|}_2^2)$, where $\Omega_i$ defines the temporal local region of a temporal step $i$ and $\widehat{T_{ik}}$ denotes the smoothed trajectory at $v_{ik}$. $w_{ij}$ is the weight used to control the temporal local smoothness of the trajectory. However, if only the smoothness is considered, there will be a trivial solution because the smoothed trajectory may be a straight line, which is over-smoothed and cannot maintain the properties of the origin trajectory, \eg, direction, tendency. The constraints that the smoothed trajectory should not be far away from the origin trajectories are applied as $L_{ts} = \sum_{i=1}^{E-1}({\| \widehat{T_{ik}} - T_{ik}\|}_2^2 + \lambda L_s)$, where $\lambda$ is introduced to balance the local smoothness term and the distance term. However, is it reasonable to require that every temporal local region be equally smooth? The local smoothness should vary dynamically according to the spatial properties of the trajectory itself and the grid, because the smoothed trajectory should not introduce additional distortions to the stabilized results. From this point of view, we further include a content preserving term in the objective function, since discarding spatial consistency can lead to distortion or content loss. The objective functions are detailed in Section~\ref{subsec:ObjectiveFunctions}.  

Specifically, to predict the dynamic weights that can be used to smooth the trajectories, we adopt a novel 3D CNN structure based on the similarity of the trajectories. As shown in Figure~\ref{fig:NetworkStructure}(c), the CNN model takes the estimated trajectories as input, which is of size $[2,E,M,N]$. The trajectories are then embedded by the transformer encoder using $1\times 1$ convolution. The embedded feature with shape $[64,E,M,N]$ are then processed by several 3D convolution layers, which then goes through a kernel decoder and a $lambda$ decoder in parallel, generating features of size $[12,E,M,N]$ and $[1,E,M,N]$. There are 12 weights estimated for each vertex since we use different weights for different dimensions and uses 3 as the temporal radius of a temporal region. The separately estimated dynamic balance term $\lambda$ and dynamic smoothness weight term $w_d$ are further substituted in the temporal smoothness objective function $L_{ts}$ to dynamic determine whether a smoothed trajectory in a temporal region should be more smooth or more close to the origin trajectory to preserve the trajectories' direction. The smoothed trajectories are then iteratively solved in the iterative smoother by solving the following equation using the estimated dynamic weights as
\begin{equation}
    {\widehat{T_{ik}}}^t = \frac{T_{ik}+\lambda_{ik} \sum_{j \in \Omega_i} w_{kij}{\widehat{T_{jk}}}^{t-1}}{1+\lambda \sum_{j \in \Omega_i} w_{kij}},
    \label{eq:jacobisolver}
\end{equation}
where $\lambda_{ik}$ and $w_{kij}$ is the output of the CNN based smoother at each specific location $k$ and temporal step $i$. In conclusion, the stabilization process can be summarized in Algorithm~\ref{alg:inference}.

\subsection{Objective Functions}\label{subsec:ObjectiveFunctions}

\subsubsection{Motion Refinement}
For the estimation of target motion vector $n_{ik}$ of each vertex $v_{ik}$, there are two constraints on it based on the temporal and spatial coherence. First, $n_{ik}$ should be close to the motion vectors of $v_{ik}$'s neighboring keypoints. Second, the target position for each keypoint $p_{ij}$ can be calculated based on its motion vector $m_{ij}$ or its projection by the homogrpahy transformation of the grid that contains $p_{ij}$, which can be derived from the four grid vertices' motion vectors \cite{liu2016meshflow,sun2010secrets}. Based on the above two constraints, we formulate the objective function for the MR module as: 
\begin{align}
\small
\label{eq:mp_objective_v}\nonumber
   L_{vm} &= \lambda_{m}\sum_{i=1}^{E-1} \sum_{k=1}^{MN} \sum_{j \in \Omega_{ik}}\left \| n_{ik} - m_{ij} \right \|_1 O_{ij} \\
   & + \lambda_{v}\sum_{i=1}^{E-1} \sum_{j=1}^{L}{\parallel p_{ij} + m_{ij} - H_{ij}  \left( p_{ij}\right)\parallel}_2^2 O_{ij},
\end{align}
where $\left \| \cdot  \right \|_1$ denotes the L1 norm, $\left \| \cdot  \right \|_2$ denotes the L2 norm, $H_{ij}$ is the homography of the grid containing $p_{ij}$, $\lambda_{m/v}$ is the weight to balance the two terms. $O_{ij}$ is the occlusion mask to alleviate the influence of dynamic objects in the motion estimation process. Denote the reconstructed motion based on grid vertices as $\widehat{OF}_i$, and the warped motion using $OF_i$ as $\omega(\widehat{OF}_i)$. The occlusion mask $O_i$ is calculated as,
\begin{equation}
    O_i = (OF_i - \omega(\widehat{OF}_i))^2 < \alpha_1 ((OF_i)^2 - (\omega(\widehat{OF}_i))^2) + \alpha_2
\end{equation}
where $\alpha_1=0.01$ and $\alpha_2=0.5$, as defined in ~\cite{sundaram2010dense,jonschkowski2020matters}. 
In addition to the keypoints-related constraints in Eq.~\eqref{eq:mp_objective_v}, we also introduce a shape preserving loss $L_{sp}$ on each grid $g_{im}$ to enhance spatial coherence, \ie,
\begin{equation}
\small
    L_{sp} = \sum_{i=1}^{E-1} \sum_{m=1}^{\left(M-1\right)\left(N-1\right)}{\left \| {\widehat{v}_{im}^3} - \left( {\widehat{v}_{im}^2} + R_{90} \left( {\widehat{v}_{im}^1} - {\widehat{v}_{im}^2} \right)\right) \right \|}_2^2,
    \label{eq:shapeloss}
\end{equation}
where ${\widehat{v}_{im}^o}=v_{im}^o+n_{im}^o, o=1,2,3,4$, $v_{im}^1 \sim v_{im}^4$ denote four vertices of $g_{im}$ in the clockwise direction, their motion vectors are $n_{im}^1 \sim n_{im}^4$, $R_{90}\left( \cdot \right)$ denotes rotation $90^\circ$ of a vector. $L_{sp}$ accounts for spatial smoothness and avoids large distortion in untextured regions. The final objective is:
\begin{equation}
    L_{MR} = L_{vm} + \lambda_{s} L_{sp},
\end{equation}
where $\lambda_{s}$ is loss weight that balances the two losses. Note that, although the above objective function can be solved using traditional solver, it takes much longer time than using our deep learning based solvers as it needs to iterate on over 1 million parameters on a standard 480p video. The idea of using DNN as a quick solver can be found in LSNet~\cite{clark2018ls} and MeshSLAM~\cite{Bloesch_2019_ICCV}.

\subsubsection{Trajectory Smoothing}\label{sec:TSObjective}
To help the CNN to predict appropriate dynamic parameters to facilitate the trajectory smoothing process, we add spatial coherence loss and content preserving loss in addition to the origin temporal smoothness loss $L_{ts}$ to characterize the desired smoothed trajectory, \ie,
\begin{equation}
     L_{TS} = L_{ts} + \lambda_s L_{sp} + \lambda_c L_{cp},
     \label{eq:ts_objective}
\end{equation}
where $L_{sp}$ is the spatial shape preserving loss similar to the one used for MR in Eq.~\eqref{eq:shapeloss} where the target position of $v_{ik}$ after smoothing is:
\begin{equation}
    \widehat{v}_{ik}=v_{ik} + \widehat{T}_{ik} - T_{ik}.
    \label{eq:targetsmoothedvertexs}
\end{equation}
Moreover, since distortion in structural areas around keypoints affects the visual experience more than those in untextured areas, we introduce a content preserving loss $L_{cp}$:
\begin{equation}
    L_{cp} = \sum_{i=1}^{E-1}\sum_{j=1}^{L}{\parallel Bili(p_{ij}) - H_{ij}(p_{ij}) \parallel}^2,
\end{equation}
where $H_{ij}$ is same as in Eq.~\eqref{eq:mp_objective_v}. Assuming $p_{ij}$ is in the grid $g_{im}$, then $Bili\left( \cdot \right)$ denotes a bilinear interpolation, \ie, $Bili(p_{ij}) = \sum_{o=1}^4 w_o \widehat{v}_{im}^o$. Here, $\widehat{v}_{im}^1\sim\widehat{v}_{im}^4$ are the target positions of four vertices in $g_{im}$ after smoothing and calculated as Eq.~\eqref{eq:targetsmoothedvertexs}. $w_1\sim w_4$ are the interpolation coefficients calculated based on the position of $p_{ij}$ relative to the four vertices of $g_{im}$ before smoothing, \ie, $v_{im}^1\sim v_{im}^4$. $L_{cp}$ indeed preserves the shape of each grid containing keypoints, which probably has structural content. Note that the proposed CNN smoother can be trained using only unstable videos based on the objective in Eq.~\eqref{eq:ts_objective}.

\begin{algorithm}
% \small
\SetAlgoLined
\KwIn{Unstable video: $\{f_i|i \in [1,E]\}$ \newline
Optical flow: $\{OF_i|i \in [1,E-1]\}$;}
\KwOut{Stabilized video: $\{\widehat{f}_i|i \in [1,E]\}$\newline
Grid motion:$\{n_{ik}|i\in [1,E-1],k\in [1,MN]\}$\newline
Estimated trajectory: $\{T_k|k \in [1,MN]\}$\newline
Smoothed trajectory: $\{\widehat{T}_k|k \in [1,MN]\}$;
}
%  \For{$i=1:E-1$}{
%   $\{p_{ij} | j \in [1,L]\} = RFNet\left(f_i\right)$;
%   $m_{ij}$ = $OF_i(p_{ij})$;\\
%  }
 \For{$i=1:E-1$}{
  $\{p_{ij} | j \in [1,L]\} = RFNet\left(f_i\right)$; $\forall j$,$m_{ij}$=$OF_i(p_{ij})$;\\
 \eIf{There are multiple planes in frame $f_i$ }
 {
 $H_i^c = MultiHomo\left(\{p_{ij}|\forall j\}, \{m_{ij}|\forall j\}\right)$; \\
 }
 {
 $H_i^c = SingleHomo\left(\{p_{ij}|\forall j\}, \{m_{ij}|\forall j\}\right)$;\\
 }
 $\Delta{m_{ij}} = p_{ij} + m_{ij} - H_i^c  \left( p_{ij}\right), \forall j \in \Lambda_i^c$;\\
 \For{$k=1:MN$}{
 $d_{ijk} = p_{ij} - v_{ik}, \forall j \in [1,L]$;\\
 $\Delta{n_{ik}} = MR\left( \{ \left( \Delta{m_{ij}}, d_{ijk} \right) | \forall j \in [1,L] \} \right)$;\\
 $n_{ik} = \Delta{n_{ik}} + \left( H_i^c  \left( v_{ik}\right) - v_{ik}\right)$;
 }
 }
 \For{$k=1:MN$}{
 $T_k = \sum_{i=1}^{E-1} n_{ik}$; $\widehat{T_k} = Smoother\left(T_k\right)$;
 }
 \For{$i=0:E$}{
 $\hat{f_i}$ = $Reprojection\left(f_i, \{T_k|\forall k\}, \{\widehat{T_k}|\forall k\}\right)$
 }
 \caption{Deep Unsupervised Trajectory based Video Stabilizer (DUT)}
 \label{alg:inference}
\end{algorithm}

\section{Experiments}\label{sec:experiments}
We evaluated our model on public benchmark dataset NUS~\cite{liu2013bundled} and compared it with state-of-the-art video stabilization methods, including deep learning based method~\cite{wang2018deep,Choi_TOG20,yu2020learning} and traditional methods~\cite{liu2011subspace,liu2016meshflow}. Both quantitative and qualitative results as well as model complexity are provided for comprehensive comparison. Ablation studies of key modules in our model were also conducted.

\subsection{Experiment Settings}
Unstable videos from DeepStab~\cite{wang2018deep} were used for training. Five categories of unstable videos from \cite{liu2013bundled} were used as the test set. The metrics introduced in \cite{liu2013bundled} were used for quantitative evaluation, including cropping ratio, distortion, and stability. Cropping ratio measures the ratio of remaining area and distortion measures the distortion level after stabilization. Stability measures how stable a video is by frequency domain analysis. All the metrics are in the range of $[0,1]$. A larger value denotes a better performance.

\begin{table}[htbp]
  \centering
  \caption{Stability scores of different methods.}
  \setlength{\tabcolsep}{0.003\linewidth}{
    \begin{tabular}{l|cccccc}
    \hline
     & {Regular} & {Parallax} & {Running} & {QuickRot} & {Crowd} & {Avg.} \\
    \hline
    Meshflow & \underline{0.843}  & 0.793  & \underline{0.839}  & 0.801  & 0.774  & 0.810  \\
    SubSpace & 0.837  & 0.760  & 0.829  & {\textbackslash{}} & 0.730  & 0.789$^*$ \\
    DIFRINT & 0.838  & 0.808  & 0.822  & \underline{0.835}  & 0.791  & \underline{0.819}  \\
    Yu \etal & \textbf{0.849} & \textbf{0.819} & 0.830 & 0.763 & \textbf{0.811} & 0.814 \\
    StabNet & 0.838 & 0.769 & 0.818 & 0.785 & 0.741 & 0.790 \\
    DUT & \underline{0.843}  & \underline{0.813}  & \textbf{0.841}  & \textbf{0.877}  & \underline{0.792}  & \textbf{0.833}  \\
    \hline
    \multicolumn{7}{p{25em}}{$^*$ The average score is somewhat higher since SubSpace fails to stabilize some videos in the category of Quick Rotation.}\\
    \end{tabular}}%
  \label{tab:StabilityScore}%
\end{table}%
\begin{table}[htbp]
  \centering
  \caption{Distortion scores of different methods.}
  \setlength{\tabcolsep}{0.003\linewidth}{
    \begin{tabular}{l|cccccc}
    \hline
     & {Regular} & {Parallax} & {Running} & {QuickRot} & {Crowd} & {Avg.} \\
    \hline
    Meshflow & 0.898  & 0.716  & 0.764  & \underline{0.763}  & 0.756  & 0.779  \\
    SubSpace & \underline{0.973}  & 0.855  & \textbf{0.939}  &{\textbackslash{}} & 0.831  & \underline{0.900}  \\
    DIFRINT & 0.934  & \underline{0.921}  & 0.873  & 0.633  & \underline{0.905}  & 0.853  \\
    Yu \etal & 0.925 & 0.836 & 0.797 & 0.221 & 0.832 & 0.722 \\
    StabNet & 0.702 & 0.573 & 0.753 & 0.574 & 0.759 & 0.672 \\
    DUT & \textbf{0.982}  & \textbf{0.949}  & \underline{0.927}  & \textbf{0.935}  & \textbf{0.955}  & \textbf{0.949}  \\
    \hline
    \end{tabular}}%
  \label{tab:DistortionScores}%
\end{table}%
\begin{table}[htbp]
  \centering
  \caption{Cropping ratios of different methods.}
  \setlength{\tabcolsep}{0.003\linewidth}{
    \begin{tabular}{l|cccccc}
    \hline
     & {Regular} & {Parallax} & {Running} & {QuickRot} & {Crowd} & {Avg.} \\
    \hline
    Meshflow & 0.686  & 0.540  & 0.584  & 0.376  & 0.552  & 0.548  \\
    SubSpace & 0.712  & 0.617  & 0.686  &{\textbackslash{}} & 0.543  & 0.639  \\
    DIFRINT & \textbf{0.922}  & \textbf{0.903}  & \textbf{0.869}  & \textbf{0.732}  & \textbf{0.882}  & \textbf{0.862}  \\
    Yu \etal & 0.858 & 0.773 & 0.649 & 0.218 & 0.824 & 0.664 \\
    StabNet & 0.537 & 0.503 & 0.512 & 0.418 & 0.497 & 0.493 \\
    DUT & \underline{0.736}  & \underline{0.709}  & \underline{0.690}  & \underline{0.673}  & \underline{0.710} & \underline{0.704}  \\
    \hline
    \end{tabular}}%
  \label{tab:CroppingRatio}%
\end{table}%
\subsection{Quantitative Results}
\begin{figure*}[htbp]
    \centering
    \includegraphics[width=\linewidth]{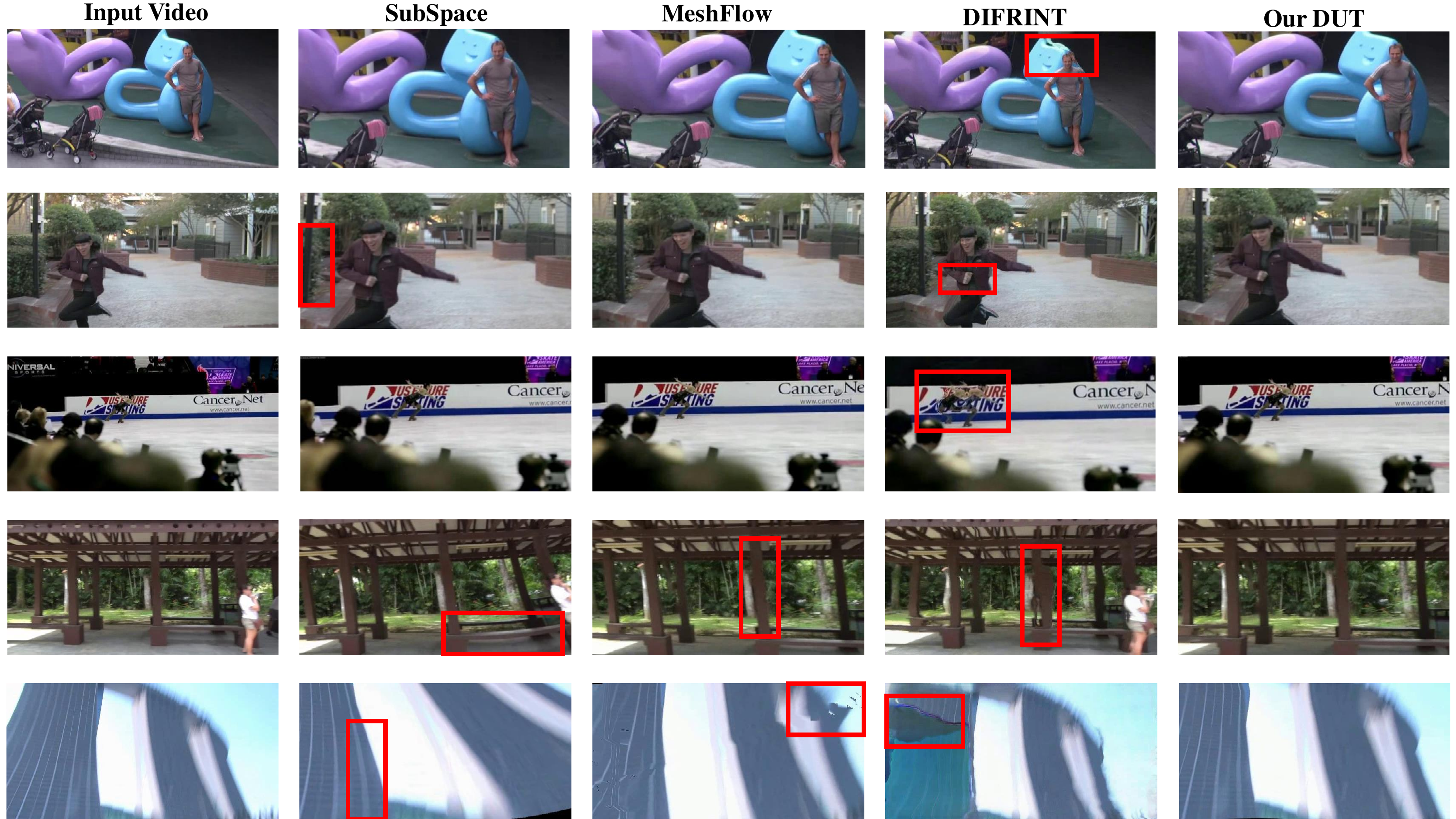}
    \caption{Visual comparison between our DUT and SubSpace~\cite{liu2011subspace}, Meshflow~\cite{liu2016meshflow}, DIFRINT~\cite{Choi_TOG20}. We present five kinds of videos for subjective evaluation, \ie, 1) moderate camera shake, 2) large dynamic objects, 3) small dynamic objects, 4) multiple planar motions, and 5) quick camera rotation and blur in the first row to the fifth row, respectively. It can be observed that our method can handle diverse scenarios compared with traditional stabilizers and deep learning based methods. More results can be found in the supplementary video.}
    \label{fig:Quality}
\end{figure*}
We compared our method with representative state-of-the-art video stabilization methods, including traditional methods: Subspace~\cite{liu2011subspace}, Meshflow~\cite{liu2016meshflow}, and deep learning methods: DIFRINT~\cite{Choi_TOG20}, Yu \etal~\cite{yu2020learning} and StabNet~\cite{wang2018deep}. 
The quantitative results are summarized in Table~\ref{tab:StabilityScore}. 
There are several empirical findings. Firstly, the proposed DUT stabilizer achieves better performance in challenging cases with multiple planar motions, \eg, in the category of Parallax. 
Secondly, the deep learning based keypoint detector empowers DUT with better capacity in dealing with challenging cases compared with other traditional stabilizers. In the category of Quick Rotation, which contains blur and large motion in the scenes, DUT still obtains a large margin over other stabilizers regarding distortion and stability metrics. Subspace~\cite{liu2011subspace} even fails for videos in this category since it is difficult to track long-term robust keypoints. 
Thirdly, compared with StabNet~\cite{wang2018deep}, DUT produces less distortion and keeps more areas after cropping. Although StabNet also uses grid based warping while being trained with paired data from the deepStab, it does not achieve a balance between stability and distortion. By contrast, DUT simply uses unstable videos for training.
As there are many possible stabilized videos that all look stable about the same, training with paired data that only provides one single stable instance for each sample may be biased. Besides, the supervisory signal from the stable video is less effective than those from the grid based pipeline, which have explicit relationships to stability, \eg, displacement derived from the grid motion and trajectory. 
Fourthly, compared with the interpolation based DIFRINT~\cite{Choi_TOG20}, DUT produces less distortion around the dynamic objects since it can handle them via robust keypoint based local grid warping. It is also interesting to find that although DIFRINT does not need cropping after stabilization, it does learn a zoom-in effect during stabilization and causes content missing problem in the stabilized results, as shown in Figure~\ref{fig:Quality}. Compared with Yu \etal~\cite{yu2020learning}, DUT does not need extra pre-stabilization step, and alleviates the limitation introduced by the pre-stabilization step, \eg, the traditional pre-stabilization makes the stabilizer does not perform well on unstable videos with quick rotation. Generally, DUT achieves the best performance regarding stability and distortion, which confirms the superiority of our trajectory based DNN stabilizer. Note that although we do not directly constrain the cropping ratio terms in the objective functions, DUT achieves impressive performance on the cropping ratio metric. We conjecture that it is because the estimated trajectory in DUT is more accurate with DNN and that the smoother part does not need to make large displacements to smooth the noise in estimated trajectories.

\subsection{Subjective Results}
Some visual results of different methods are shown in Figure~\ref{fig:Quality}. For videos containing dynamic objects (the 1st-3rd rows), DIFRINT~\cite{Choi_TOG20} is prone to produce ghost artifacts around dynamic objects due to the neighboring frame interpolation, which is unaware of dynamic foregrounds. Subspace~\cite{liu2011subspace} may generate some distortion if the dynamic object is large as in the 2nd row. For scenes containing parallax and occlusion (the 4th row), Subspace also produces distortion due to the inaccurate feature trajectory. DIFRINT generates hollows since the interpolation cannot hallucinate the occluded objects. Meshflow~\cite{liu2016meshflow} produces shear artifacts due to the inaccurate homography estimated from a single plane. DUT has no such drawbacks since it adopts an adaptive multi-homography estimation method. For videos containing quick camera rotation and large blur (the 5th row), Subspace and Meshflow produce strong distortion due to the inaccurate trajectory estimation based on hand-crafted features. There is a dark patch in DIFRINT's result, which is caused by interpolating far away regions from adjacent frames to the inaccurate area of the current frame due to the quick camera motion. In conclusion, DUT is robust to different camera motion patterns and diverse scenes, owing to its powerful representation capacity of DNNs.

\subsection{Ablation Studies}
As we tackle the learning of video stabilization by a divide-and-conquer strategy, \ie, learning to estimate accurate enough trajectory and learning to smooth the trajectories with adaptative kernels, we will do ablation studies related to the two subtasks separately and discuss the results as follows.

\subsubsection{Motion Refinement}
\label{sec:MotionRefineAB}
{
Motion refinement is crucial for video stabilizers to provide accurate trajectory. To better investigate the influence of the Motion Refinement Module, we consider warping the neighboring frames to the current frame based on the estimated grid motions and calculate the error based on the warped frames and the current frame, since we do not have ground truth grid motion to directly assess the quality of motion estimation. Specifically, we consider two specific metrics for evaluation. First, if the grid motion is accurate, the warped frames should be close to the ground truth frame, \ie, there should be an identity homography between them. Therefore, we can calculate the Frobenius distance between the estimated homography and the identity homography as the first metric, \ie, “distance” in Table~\ref{tab:MotionAblation}. Second, following \cite{liu2013bundled}, we calculate the ratio between the largest two eigenvalues of the homography as the distortion metric, \ie, “distortion” in Table~\ref{tab:MotionAblation}.

\begin{table}[htbp]
  \centering
  \caption{{The evaluation results of different motion refinement strategies. Please see Section~\ref{sec:MotionRefineAB} for the definition of the metrics.}}
    \begin{tabular}{l|cc}
    \hline
          & MF & MR \\
    \hline
    Distance$\downarrow$& 0.069 & 0.043 \\
    Distortion$\uparrow$& 0.941 & 0.956 \\
    \hline
    \end{tabular}
    %}%
  \label{tab:MotionAblation}%
\end{table}%

\begin{table}[htbp]
  \centering
  \caption{{The evaluation results of different motion refinement strategies. Please see Section~\ref{sec:MotionRefineAB} for more details.}}
    \begin{tabular}{l|ccc}
    \hline
          & Cropping & Distortion & Stability \\
    \hline
    MF    & 0.752  & 0.846  & 0.781  \\
    MR    & 0.771  & 0.894  & 0.804  \\
    \hline
    \end{tabular}%
  \label{tab:MFMRCompare}%
\end{table}%

The results are summarized in Table~\ref{tab:MotionAblation}. MF denotes using the median filter for motion refinement where the initial motion and keypoints are generated by the proposed motion initialization module, while MR denotes using the proposed module for motion refinement. As can be seen, MR achieves better scores than MF in terms of both metrics, showing that the trajectory estimated by the proposed motion refinement module is more accurate than that using the median filter, which is attributed to the dynamic selection of reliable keypoints for motion refinement. In addition, based on the estimated trajectories in the above two settings, we use the same smoother (\ie, the proposed trajectory smoothing module) to obtain the stabilization results. The results are presented in Table~\ref{tab:MFMRCompare}. It can be seen that using the trajectories estimated by the MR module leads to better stabilized results, which further validates the effectiveness of the proposed motion refinement module in obtaining accurate motion and trajectories.

}

\begin{figure}[htbp]
    \centering
    \includegraphics[width=\linewidth]{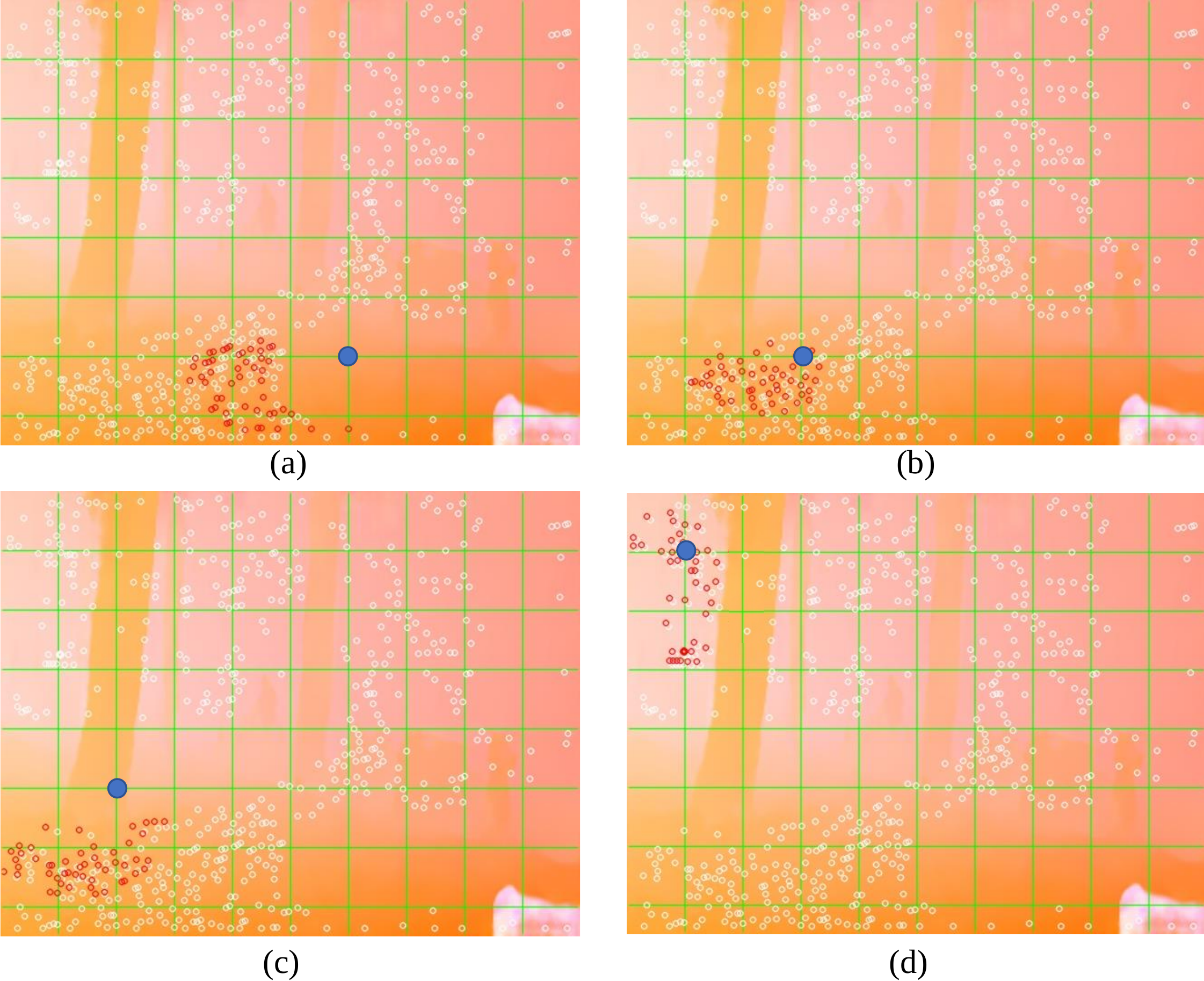}
    \caption{Visualization of the important supporting keypoints of different grid vertices, which have large weights in our MR module. For the grid vertex where there are few keypoints in its neighborhood, \eg, (a), and (c), our motion propagation module tends to rely on those keypoints which have similar motions to the grid vertex. For the grid vertex which is surrounded by rich keypoints, \eg, (b), and (d), our motion propagation module relies more on the nearby keypoints which have similar motions to the grid vertex.}
    \label{fig:MotionAttention}
\end{figure}

To further demonstrate the effectiveness of our MR stage, we visualized the important supporting keypoints for some grid vertices selected by the MR stage in Figure~\ref{fig:MotionAttention}. Note that we sorted the weights of all the keypoints generated by our MR stage for each grid vertex, and selected those with large weights as the supporting keypoints as shown in red circles in Figure~\ref{fig:MotionAttention}. Other keypoints are marked as white circles. The grid vertex is marked as the blue point. For the grid vertex where there are few keypoints in its neighborhood, \eg, (a) and (c), the MR stage prefers choosing those keypoints which have similar motions to the grid vertex. For the grid vertex which is surrounded by rich keypoints, \eg, (b) and (d), the MR stage relies more on the keypoints which are nearby and have similar motions to the grid vertex. Compared with the median filter which treats each point in its neighborhood equally, our MR module can adaptively choose the supporting keypoints from all the keypoints for each grid vertex according to their distances and motion patterns.

To investigate the impact of different terms in the objective function, we carried out an empirical study of the hyper-parameters, \ie, the loss weights of the L1 loss term $L_1$ and the distortion constraint term $L_d$. The objective function is reformulated as:
\begin{equation}
    L_{MR} = \lambda_{1}L_1 + \lambda_{2}L_{d},
\end{equation}
\begin{equation}
    L_{1}=\sum_{i=1}^{E-1} \sum_{k=1}^{MN} \sum_{j \in \Omega_{ik}}\left \| n_{ik} - m_{ij} \right \|_1 O_{ij},
\end{equation}
\begin{equation}
    L_{d} = \lambda_{v}\sum_{i=1}^{E-1} \sum_{j=1}^{L}{\parallel p_{ij} + m_{ij} - H_{ij}  \left( p_{ij}\right)\parallel}_2^2 O_{ij} + \lambda_{s} L_{sp}.
\end{equation}

\begin{figure}[htbp]
    \centering
    \includegraphics[width=\linewidth]{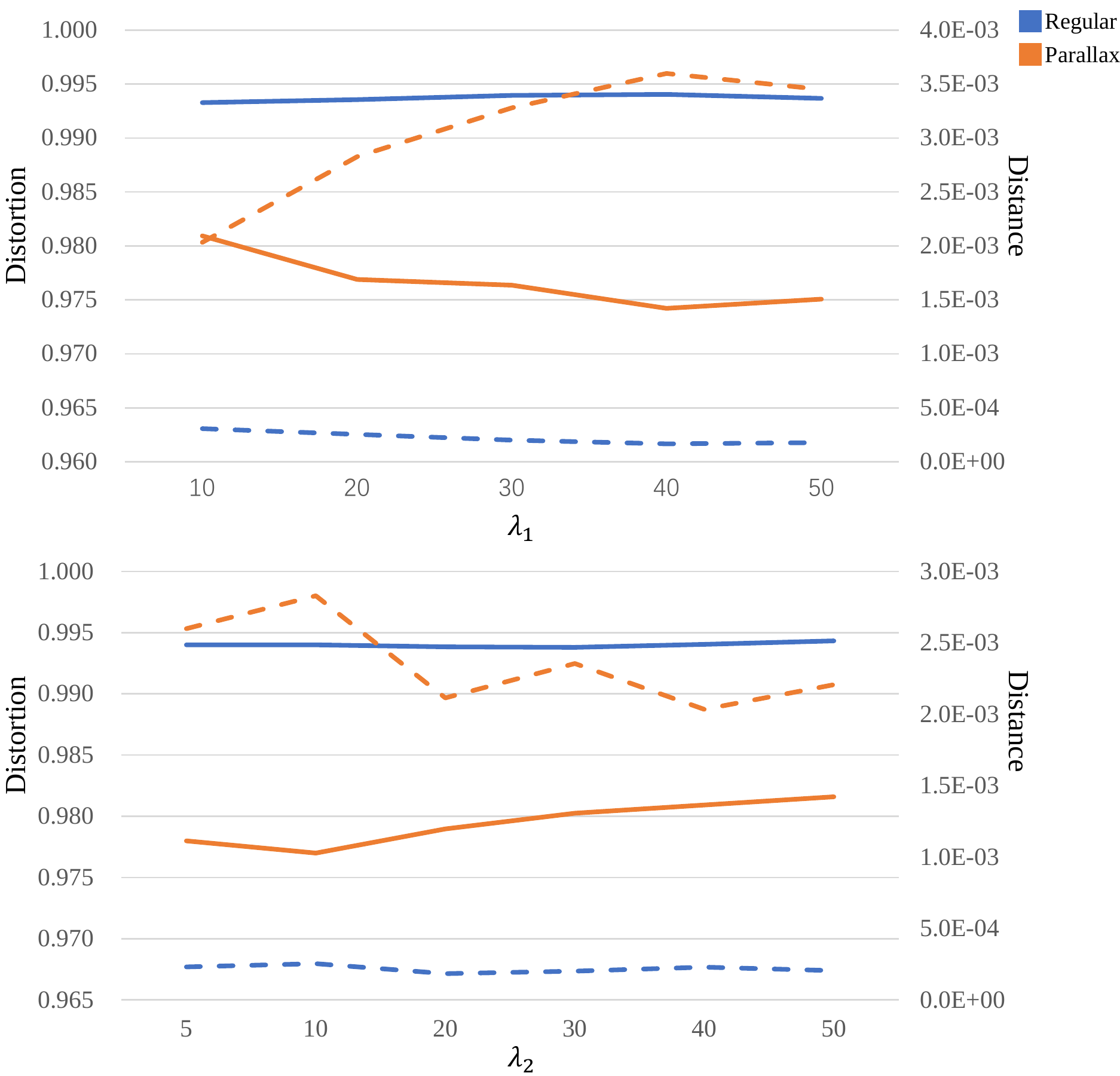}
    \caption{Empirical study of the loss weights in the objective function of the MR module. The solid lines represent the distortion metric and the dashed lines represent the distance metric.}
    \label{fig:MPLambda1}
\end{figure}

We changed the weights of $\lambda_1$ and $\lambda_2$ and evaluated the MR module on test videos from the regular and parallax categories. Note that $\lambda_{v}$ and $\lambda_{s}$ were kept 1:1 according to their amplitudes on a subset of training data. The results are plotted in Figure~\ref{fig:MPLambda1}. As can be seen, with an increase of $\lambda_1$, both the distortion and distance metrics become marginally better for the regular category, while they become worse for the parallax category. Using a large $\lambda_1$ to emphasize $L_1$ in the objective function, the MR model acts like a median filter, which is not able to deal with multiple planar motions in the videos from the parallax category. By contrast, with an increase of $\lambda_d$, our MR model achieves better performance for the parallax category in terms of both distortion and distance metrics. Generally, the MR module is robust to different hyper-parameter settings for the regular category. We set $\lambda_1$ as 10 and $\lambda_2$ as 40 in other experiments, \ie,$\lambda_m = 10$, $\lambda_v = \lambda_s = 40$.

\subsubsection{Trajectory Smoothing}

\begin{figure}[htbp]
    \centering
    \includegraphics[width=0.8\linewidth]{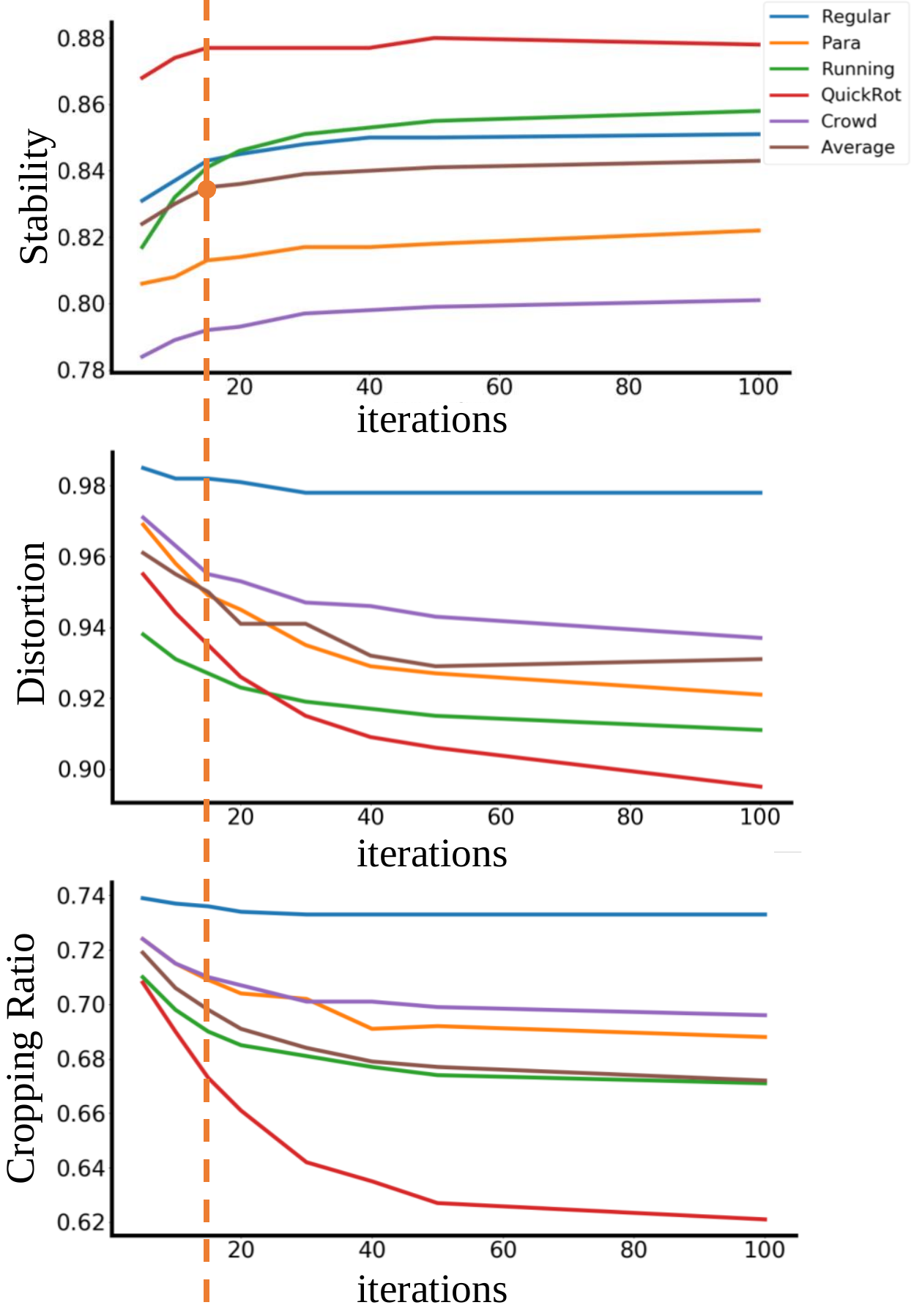}
    \caption{The stability, distortion, and cropping ratio at different settings of smoothing iterations. }
    \label{fig:DistortionMeasure}
\end{figure}

To provide a thorough analysis of the hyper parameters used in the trajectory smoothing stage, we separately investigate the influence of iterations and weights in the trajectory smoothing stage's objective functions regards the stability, distortion and cropping ratio. To evaluate the influence of different iterations of the TS module, we ran our smoother with different iterations, \ie, 5, 10, 15, 20, 30, 40, 50, 100 and plotted the average stability and distortion scores as well as the cropping ratios based on the NUS dataset in Figure~\ref{fig:DistortionMeasure}.

\begin{figure}[htbp]
    \centering
    \includegraphics[width=\linewidth]{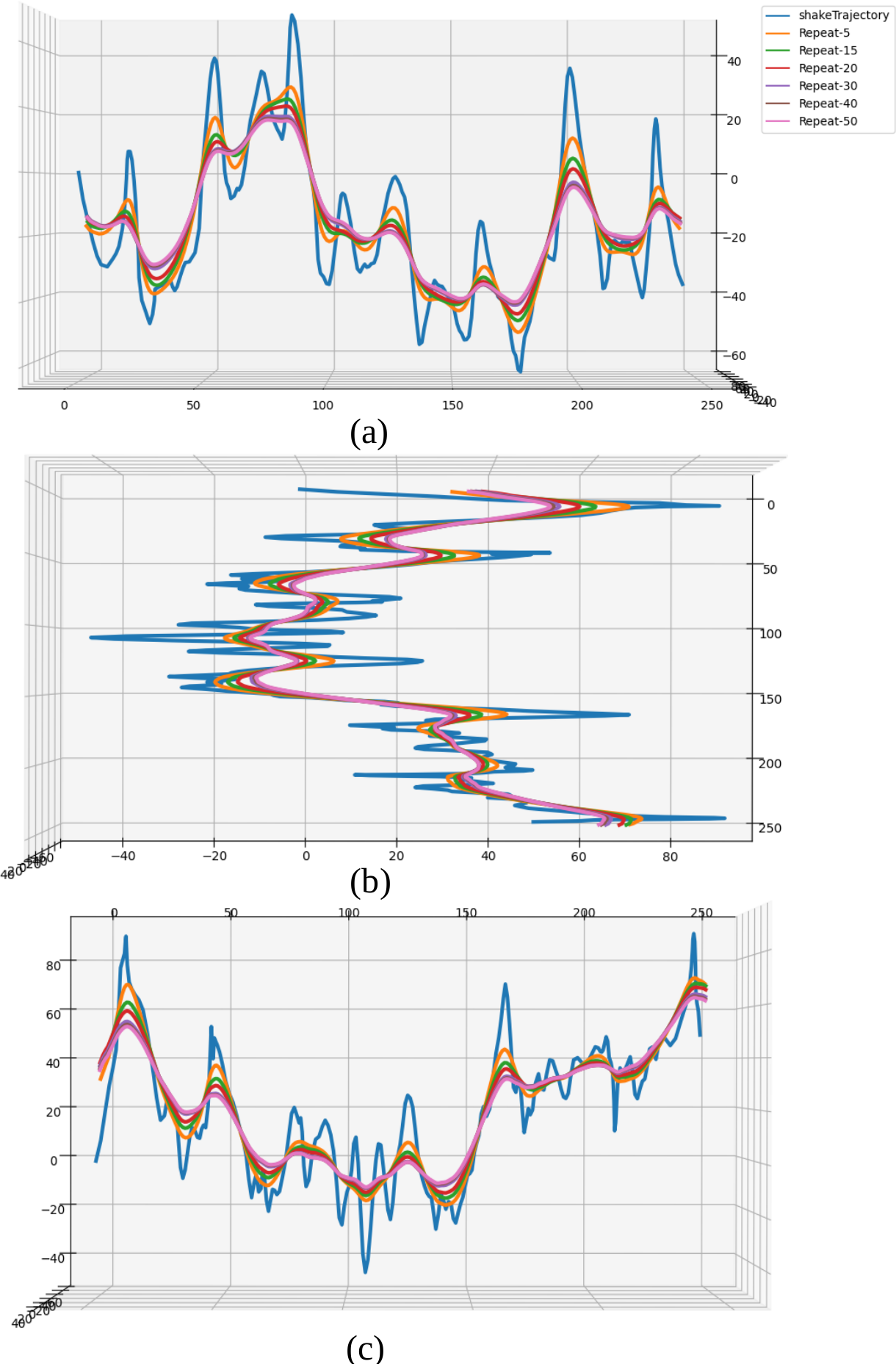}
    \caption{Visualization of the smoothed trajectories using different smoothing iterations. (a) The Front view. (b) The profile view. (c) The top view. The unstable trajectory is shown in blue and smoothed trajectories are shown in different colors.}
    \label{fig:RepeatativeTrajectory}
\end{figure}

It can be seen that the stability score increases rapidly at the beginning and then saturate after 15 iterations for all categories. For the distortion and cropping ratio, they decrease fast at the beginning and then decrease slowly. Besides, we plotted the smoothed trajectories using different smoothing iterations for a video from the Regular category in Figure~\ref{fig:RepeatativeTrajectory}, where the front view, profile view, and top view of the trajectories are plotted in (a)-(c), respectively. The unstable trajectory is shown in blue and smoothed trajectories are shown in different colors. As can be seen, with the increase of the iterations, the trajectory becomes more smoothing and deviates from the original trajectory rapidly at the beginning. Then, it changes slowly and converges after a few iterations. Generally, for different categories, there are slightly different critical points that can make a trade-off between these metrics. We choose 15 iterations for all categories according to the average performance to make a trade-off between different metrics as well as keep the computational efficiency of our method.

The impact of different terms in the objective function of TS module is also investigated. We carried out an empirical study of the loss weights. The objective function was reformulated as:
\begin{equation}
\small
    L_{TS}=\sum_{i=1}^{E-1}({\| \widehat{T_{ik}} - T_{ik}\|}_2^2 + \lambda_1 L_{smooth} + \lambda_2 L_{distortion},
\end{equation}
\begin{equation}
\small
    L_{smooth} = \sum_{j\in\Omega_i} w_{ij}{\| \widehat{T_{ik}} - \widehat{T_{jk}} \|}_2^2),
\end{equation}
\begin{equation}
\small
    L_{distortion} = \lambda_s L_{sp} + \lambda_c L_{cp}.
\end{equation}

\begin{figure}[htbp]
    \centering
    \includegraphics[width=\linewidth]{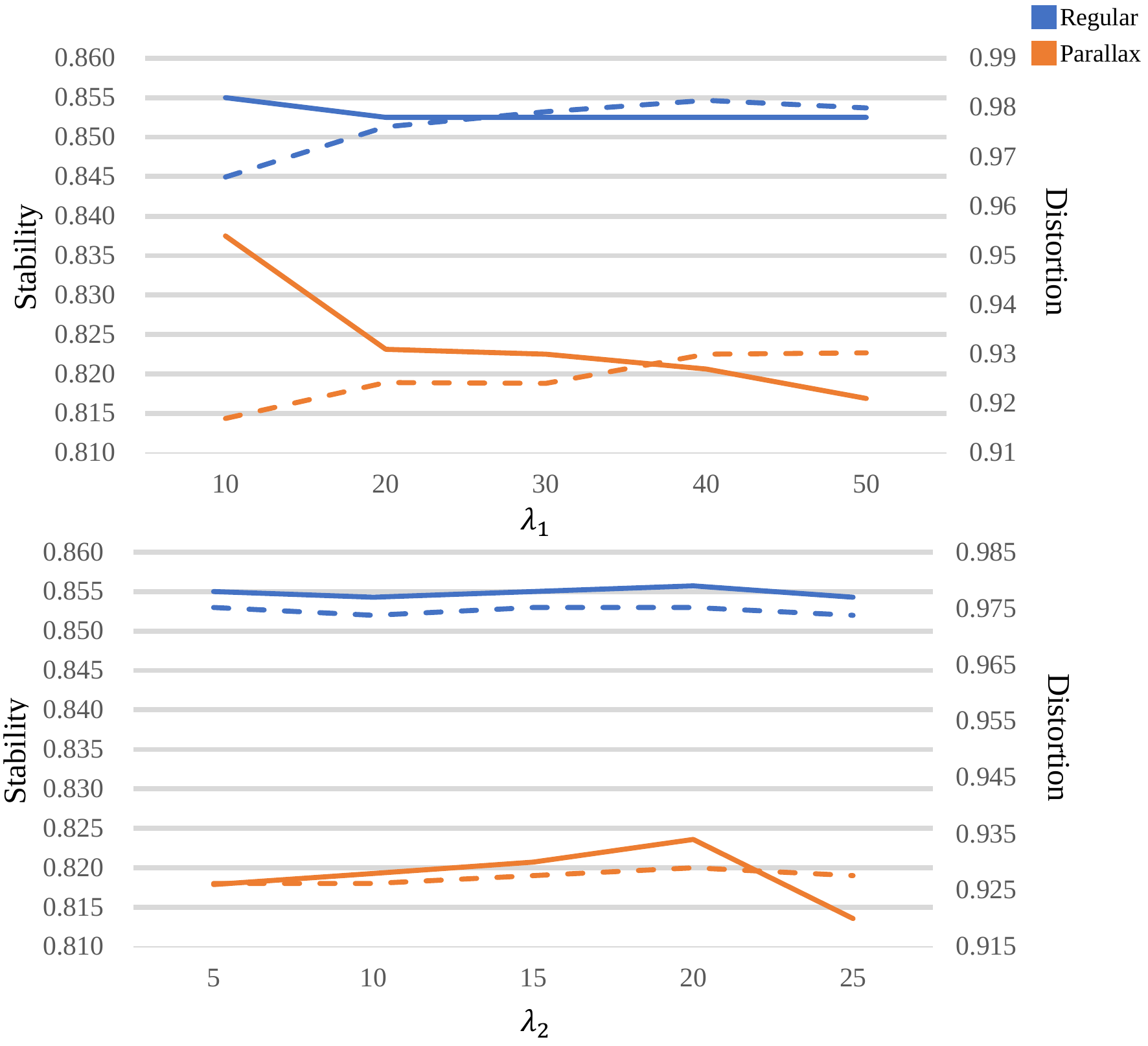}
    \caption{Empirical study of the loss weights in the objective function of the TS module. The solid lines represent the distortion metric and the dashed lines represent the stability metric.}
    \label{fig:TSLambda1}
\end{figure}

We changed the weights of $\lambda_1$ and $\lambda_2$ and evaluated the TS module on test videos from the regular and parallax categories. Note that $\lambda_{s}$ and $\lambda_{c}$ were kept 2:1 according to their amplitudes on a subset of training data. The results are plotted in Figure~\ref{fig:TSLambda1}. As can be seen, with the increase of $\lambda_1$, more stable results can be achieved for both categories while the distortion scores decrease, \ie, marginally for the Regular category but significantly for the Parallax category. It is reasonable since the distortion term contributes less to the objective function. When varying $\lambda_2$, the stability score does not change a lot while the distortion scores first increase and then decrease at the late stage, especially for the Parallax category. The distortion term encourages the transformation in each grid to keep the grid shape. However, the grid around the boundaries between two planes may not be necessary to keep its shape due to the different planar motions. Thereby, a large distortion term will lead to distortions in these regions.  Generally, the TS module is robust to different settings for the regular category. We set $\lambda_1$ as 15 and $\lambda_2$ as 20 in other experiments, \ie, $\lambda = 15$, $\lambda_s = 40$, and $\lambda_c = 20$.

{

\begin{table}[htbp]
  \centering
  \caption{{The evaluation results of different smoothers. TR represents the traditional smoother while TS denotes the proposed trajectory smoothing module.}}
    \begin{tabular}{l|cccc}
    \hline
          & Cropping & Distortion & Stability & Iterations \\
    \hline
    TR    & 0.771  & 0.878  & 0.804 & 100  \\
    TS    & 0.771  & 0.894  & 0.804 & 15 \\
    \hline
    \end{tabular}%
  \label{tab:SmootherComapre}%
\end{table}%

\begin{figure}
    \centering
    \includegraphics[width=\linewidth]{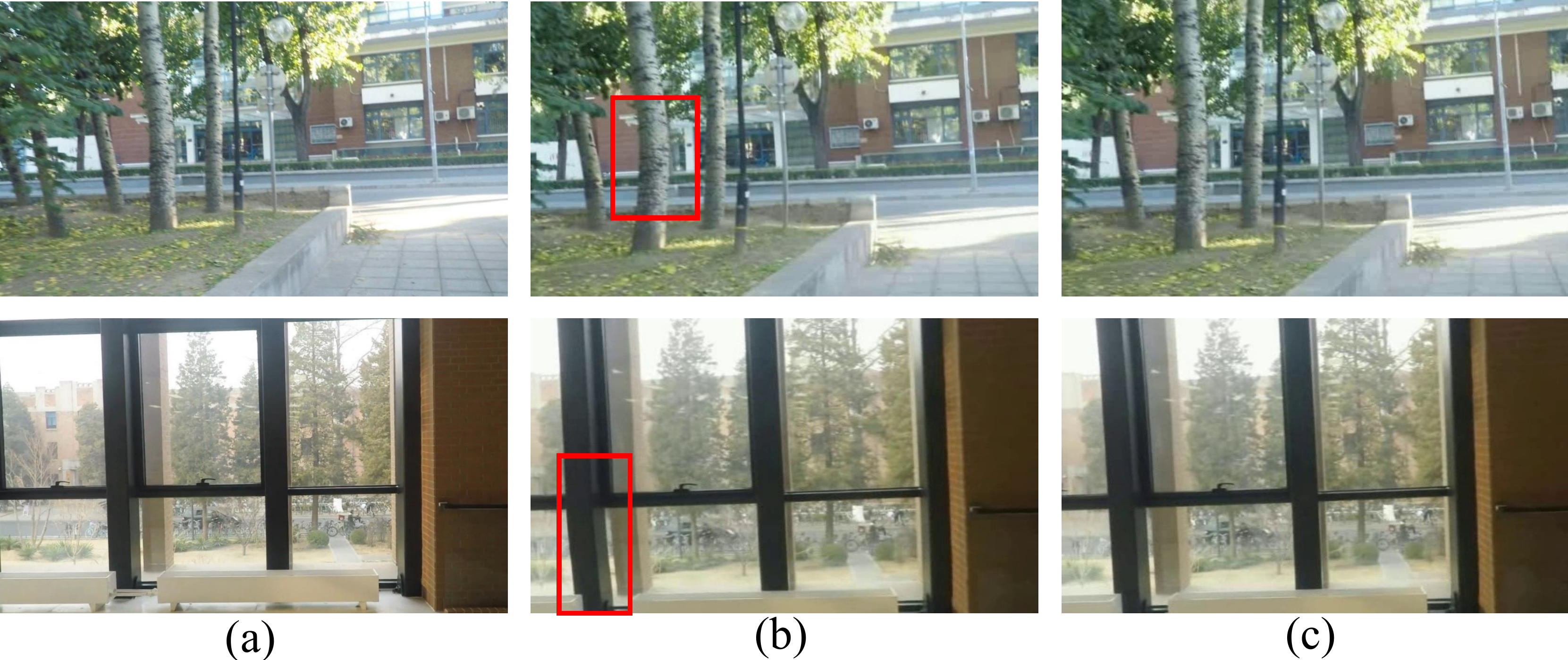}
    \caption{{Qualitative comparison of different smoothers. (a) Input frames. (b) The stabilized results using traditional smoother. (c) The stabilized results using the proposed TS module. The red boxes indicate the regions with distortion.}}
    \label{fig:TSCompare}
\end{figure}

To further validate the effectiveness of the proposed trajectory smoothing module, we use the trajectories estimated by the MR module as input and apply both the proposed trajectory smoothing module and the traditional smoother [11] for stabilization. We iteratively smooth the trajectories using each smoother until obtaining similar stability scores. The total iterations are recorded, while the distortion scores and cropping ratios of stabilized videos are calculated. The results are presented in Table~\ref{tab:SmootherComapre} and Figure~\ref{fig:TSCompare}, where TR represents the traditional smoother while TS represents the proposed trajectory smoothing module. As can be seen, the proposed trajectory smoothing module uses fewer iterations to get similar stability scores, \ie, 15 v.s. 100, showing that the proposed TS module is more computationally efficient than the traditional counterpart. Besides, TS module gets higher distortion scores, demonstrating the superiority of the dynamic smoothing kernels in the proposed TS module. 

\subsubsection{Motion Refinement and Trajectory Smoothing}

\begin{table}[htbp]
  \centering
  \caption{{The comparison between different pipelines. MeshFlow~\cite{liu2016meshflow}+MI represents using the proposed motion initialization module for motion initialization, while keeping the median filter for motion refinement and the traditional smoother for trajectory smoothing.}}
    \begin{tabular}{l|ccc}
    \hline
          & Cropping & Distortion & Stability \\
    \hline
    MeshFlow~\cite{liu2016meshflow}+MI & 0.756  & 0.846  & 0.767  \\
    DUT   & 0.771  & 0.894  & 0.804  \\
    \hline
    \end{tabular}%
  \label{tab:pipelineCompare}%
\end{table}%

To investigate the effectiveness of the proposed MR and TS modules, we use the same motion initialization module as ours in MeshFlow~\cite{liu2016meshflow} to isolate the influence of motion estimation and compare it with our DUT. The results are presented in Table~\ref{tab:pipelineCompare}, where MeshFlow~\cite{liu2016meshflow}+MI represents using the proposed motion initialization module for motion initialization, while keeping the median filter for motion refinement and the traditional smoother for trajectory smoothing. It can be seen that using the proposed MR and TS modules leads to better performance in terms of distortion and stability since the MR module can estimate more accurate trajectories while the TS module uses dynamic weights that can adapt to different trajectories.

}

\subsection{Model Complexity Comparison}

\begin{table}[htbp]
  \centering
  \caption{Model Complexity of Different Methods.}
    \setlength{\tabcolsep}{0.02\linewidth}{\begin{tabular}{l|cccc}
    \hline
          & Subspace~\cite{liu2011subspace} & StabNet~\cite{wang2018deep} & DIFRINT~\cite{Choi_TOG20} & DUT \\
    \hline
    Per-frame (ms) & 140 & \textbf{58} & 364 & 71 \\
    Params (M) & \textbackslash{} & 30.39 & 9.94 & \textbf{0.63} \\
    FLOPs (G) & \textbackslash{} & 47.55 & 67.78 & \textbf{30.42} \\
    \hline
    \end{tabular}}%
  \label{tab:TimeCost}%
\end{table}%
The running time, number of parameters, and number of computations of both traditional and deep learning methods are summarized in Table~\ref{tab:TimeCost}. For a fair comparison, we run Subspace~\cite{liu2011subspace}, StabNet~\cite{wang2018deep}, DIFRINT~\cite{Choi_TOG20} and our DUT on a 640*480 video with 247 frames for 20 times and calculate the average per frame time cost. DUT has fewer parameters and computations, outperforming all other methods. Besides, it is much faster than Subspace and DIFRINT while being only slightly slower than StabNet.

\subsection{Robustness Evaluation}

\begin{table}[htbp]
  \centering
  \caption{Robustness evaluation of the proposed DUT with respect to different levels and types of noise. G: Gaussian noise; SP: Salt and Pepper noise.}
    \setlength{\tabcolsep}{0.02\linewidth}{
    \begin{tabular}{c|ccc}
    \hline
          & \multicolumn{1}{l}{Stability$\uparrow$} & \multicolumn{1}{l}{Distortion$\uparrow$} & \multicolumn{1}{l}{Cropping$\uparrow$} \\
    \hline
    No-Noise & 0.833 & 0.949 & 0.704 \\
    G-5 & 0.832 & 0.949 & 0.704 \\
    G-10 & 0.832 & 0.949 & 0.704 \\
    G-15 & 0.832 & 0.949 & 0.704 \\
    G-20 & 0.832 & 0.949 & 0.704 \\
    SP & 0.830 & 0.948 & 0.704 \\
    Blank & 0.831 & 0.949 & 0.704 \\
    \hline
    \end{tabular}}%
  \label{tab:DUTRobust}%
\end{table}%

The optical flow calculated by traditional or deep learning-based methods may contain errors in some areas, \eg dynamic objects or untextured regions. To evaluate the robustness of our DUT concerning inaccurate optical flow values, we randomly added noise on the optical flow map and ran our stabilizer based on it accordingly. Three types of noise were included in the experiments, \ie, Gaussian noise (denoting ``G''), Salt and Pepper noise (denoting ``SP''), and the value missing error (denoting ``Blank''). Specifically, we added different types of noise on 10$\%$ regions randomly chosen from each optical flow map. We set four different levels of standard deviations for the Gaussian noise, \ie, 5$\%$ (G-5), 10$\%$ (G-10), 15$\%$ (G-15), and 20$\%$ (G-20) of the maximum flow values. We ran the experiment three times for each setting and calculated the average metrics. The results are summarized in Table~\ref{tab:DUTRobust}. As can be seen, our model is robust to various types of noise and different noise levels, \ie, only a marginal performance drop is observed when there is noise in the optical flow map. It can be explained as follows. First, the detected keypoints are always distinct feature points that are easy to get accurate optical flow. Second, the keypoints are distributed sparsely on the image that some of them may not be affected by the randomly added noise. Third, our CNN-based motion propagation module can deal with the noise since it leverages an attention mechanism to adaptively select important keypoints by assigning high weights and aggregate their motions in a weighted sum manner. Thereby, it can be seen as an adaptive filter to deal with the noise (\ie, inaccurate optical flow values).

\subsection{User Study}

\begin{figure}[htbp]
    \centering
    \includegraphics[width=\linewidth]{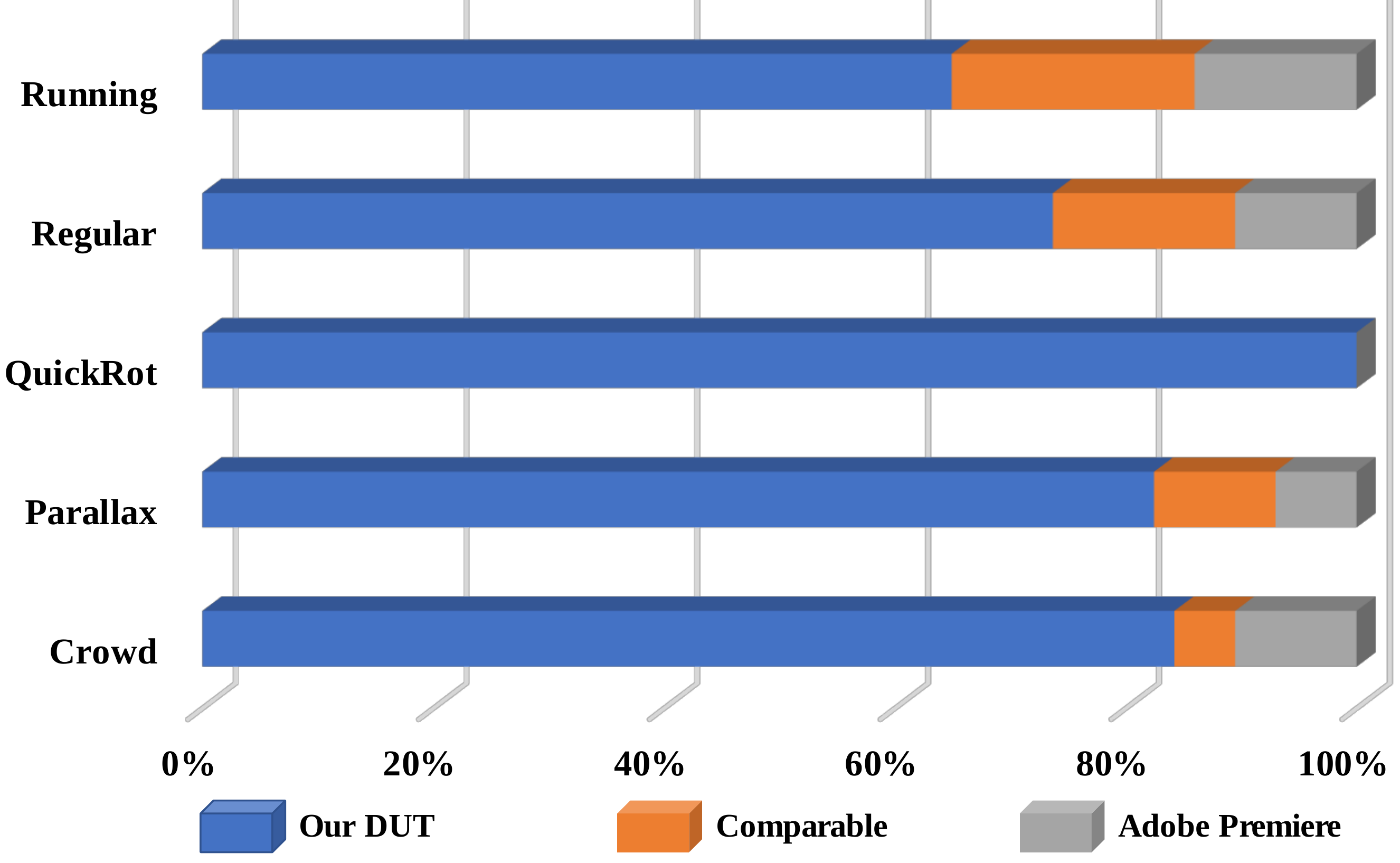}
    \caption{User preference study.}
    \label{fig:UserStudy}
\end{figure}

\begin{figure*}[htbp]
    \centering
    \includegraphics[width=0.9\linewidth]{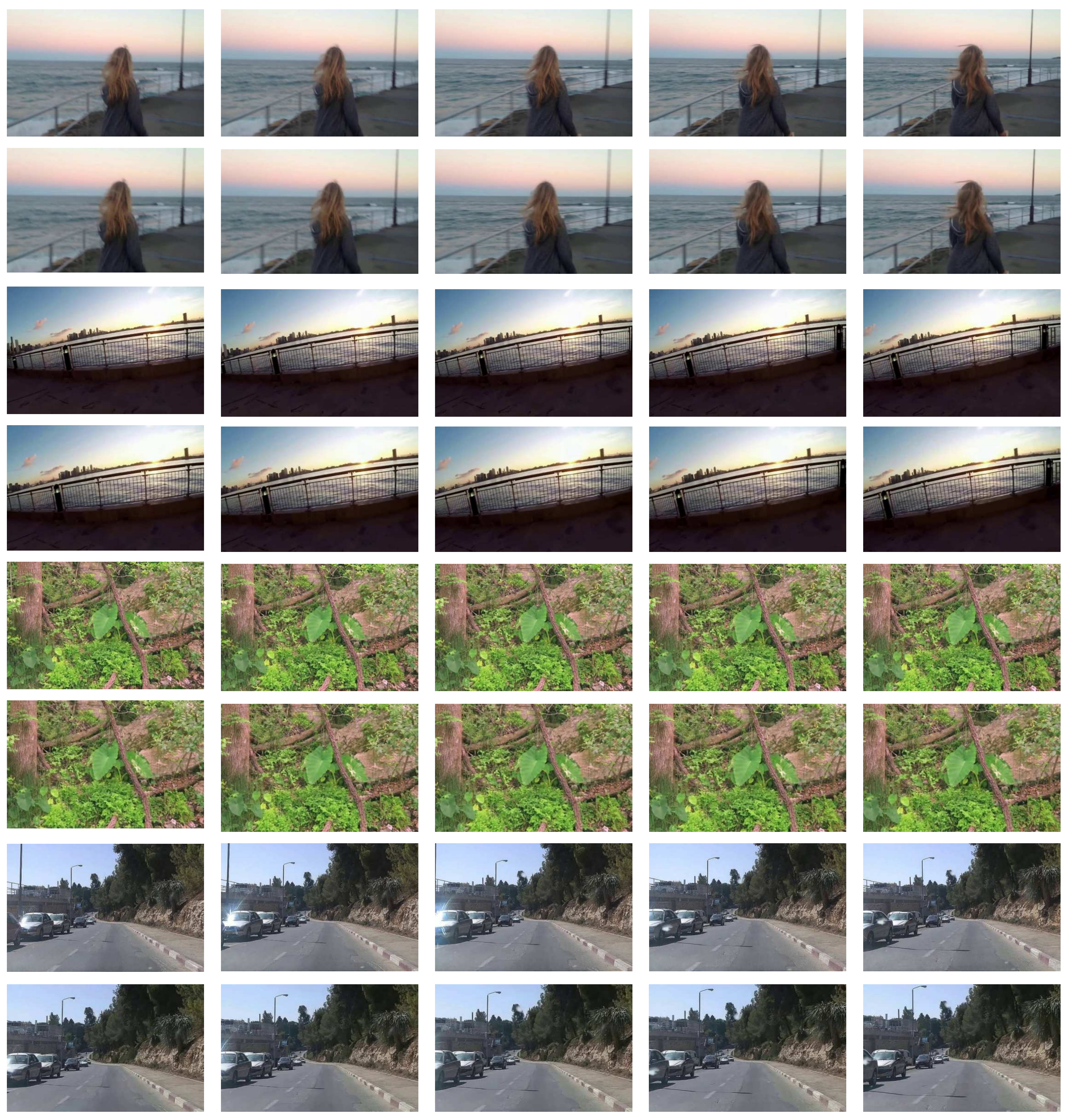}
    \caption{{Some stabilization results of the DUT on videos from different scenes regarding lighting conditions and camera motions. The odd-numbered rows are the input frames, while the even-numbered rows are the stabilized results of DUT. The first and second video clips are taken on cloudy days and at nightfall, respectively, while the third video clip is taken during climbing, and the fourth video clip is taken during driving. More results are shown in the supplementary video.}}
    \label{fig:YoutubeDemo}
\end{figure*}

Evaluation of video stabilization is also a subjective task since different subjects may have different visual experience and tolerance on stability, distortion, and cropping ratio. Therefore, we carried out a user study to compare our model with the commercial software Adobe Premiere Pro CC 2019 for video stabilization. Note that, to the best of our knowledge, Adobe Premiere and Adobe After Effects adopt the same stabilizer with the default setting. First, we chose three representative videos from each of the five categories in the NUS~\cite{liu2013bundled} dataset to constitute the test videos. 25 subjects participated in our study with ages from 18 to 30. Each subject was aware of the concept of stability, distortion, and cropping ratio after a pre-training phase given some test videos and results by other methods. Then, the stabilized videos generated by our DUT and Adobe Premiere were displayed side-by-side but in random order. Each subject was required to indicate its preference according to the stabilized video quality in terms of the aforementioned three aspects. The final statistics of the user study are summarized in Figure~\ref{fig:UserStudy}. It can be seen that most users prefer our stabilization results than those by Adobe Premiere. We find that since Adobe Premiere uses traditional keypoints detectors, the detected keypoints in consecutive frames may drift, especially in the videos from the quick rotation category, leading to less satisfying stabilization results than our deep learning-based model.

{
\subsection{Stabilization results on more scenes}
\label{sec:MoreScene}

\begin{table}[htbp]
  \centering
  \caption{{Qualitative results of representative stabilizers and DUT on videos from different scenes regarding lighting conditions and camera motions. MeshFlow~\cite{liu2016meshflow}+MI represents using the proposed motion initialization module for motion estimation while using traditional motion refinement and trajectory smoother for stabilizing.}}
    \begin{tabular}{l|ccc}
    \hline
          & Cropping & Distortion & Stability \\
    \hline
    StabNet~\cite{wang2018deep} & 0.537 & 0.672 & 0.780 \\
    MeshFlow~\cite{liu2016meshflow}+MI & 0.685 & 0.768 & 0.842 \\
    DUT   & 0.734 & 0.822 & 0.854 \\
    \hline
    \end{tabular}%
  \label{tab:YoutubeDemo}%
\end{table}%

To further evaluate the generalization ability of the proposed DUT stabilizer, we collect different unstable videos from YouTube, which contain various camera motions, such as walking, running, driving, and climbing, and cover different light conditions. Different stabilizers and our DUT are evaluated on these videos. Both the quantitative and qualitative results are provided in Table~\ref{tab:YoutubeDemo} and Figure~\ref{fig:YoutubeDemo}. It can be observed that the proposed DUT stabilizer still outperforms previous stabilizers on these videos, \eg, it achieves a gain of 0.07 stability score and 0.15 distortion score over the popular warping-based stabilizer StabNet~\cite{wang2018deep}. These results confirm the generalization ability of DUT regarding different camera motions and light conditions, owing to its powerful representation ability\footnote{{More results of videos containing challenging scenarios, \eg, textureless regions, different camera motions, various light conditions, are available in the \href{https://drive.google.com/file/d/1keQ8SLFRlwtehiI7ezUEA0GRV7rfhSHf/}{supplementary video}}}.
}

\subsection{Limitations and Discussion}\label{sec:limitation}

{
DUT decomposes the video stabilization task into learning to estimate accurate motion trajectories and smooth them for video stabilization in an unsupervised manner. Although the pipeline is similar to traditional stabilizers, leveraging deep neural networks yields robust and fast stabilization results. However, there are some hyper-parameters that can be further carefully tuned, \eg, the number of smoothing iterations for different types of videos and the multi-planar thresholds in the motion refinement module. For example, although the DUT with default parameters generally performs well, it may cause distortion on some videos, as shown in Figure~\ref{fig:Failure}. To address this issue, more research efforts on devising efficient strategy for adaptive hyper-parameter setting or leveraging the attention mechanism for a more robust solution, can be made in the future work. Besides, since the satisfactory video stabilization result of a shaky video is not unique, how to objectively evaluate the performance of video stabilizers by either full-reference quality assessment \cite{seshadrinathan2009motion} or non-reference quality assessment \cite{zhang2018intrinsic} remains an open challenge.

\begin{figure}
    \centering
    \includegraphics[width=8cm]{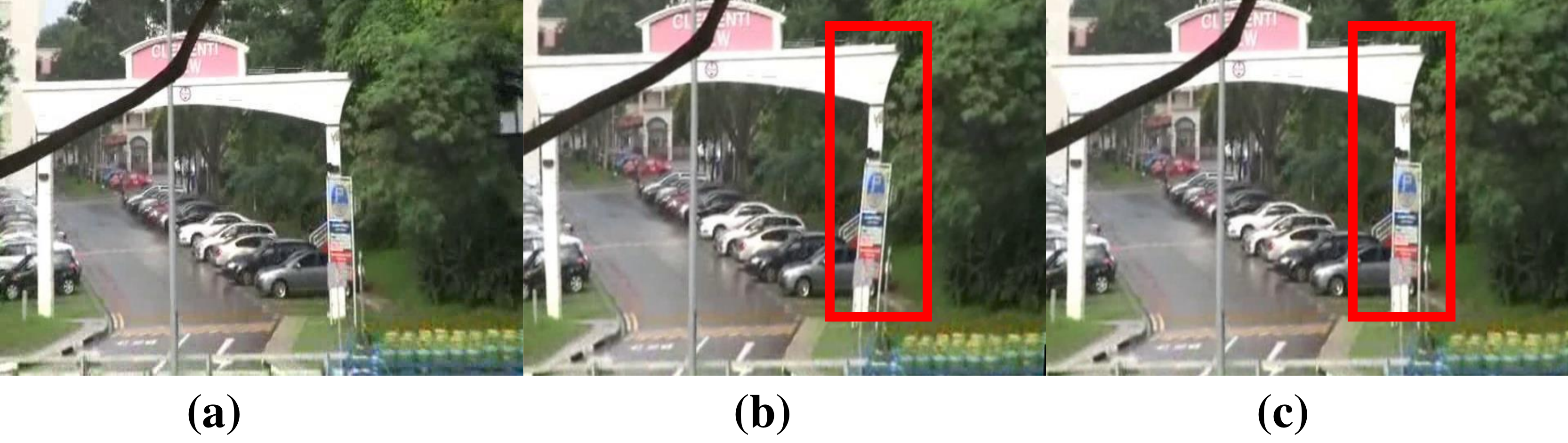}
    \caption{
    {Failure case of DUT. (a) A test video containing multiple planar motion, \eg, gate and road. (b) Failure result by DUT due to inappropriate setting of the multi-planar threshold. (c) Result of DUT with carefully tuned threshold.}
    }
    \label{fig:Failure}
\end{figure}

It is noteworthy that in this paper we focus on improving the performance of warping-based video stabilization methods, especially regarding stability and distortion of their stabilized videos. To this end, we propose the deep neural network-based DUT model as well as explore an unsupervised training scheme. Nevertheless, it also suffers from the low crop ratio issue like other warping-based methods due to content loss. The reason is that the current frame may not have the candidate pixels for unstable to stable warping, \ie, resulting in a blank area that should be cropped in the final stabilized video for a better visual experience. Inspired by the success of DIFRINT, we believe that such pixels can be traced from neighboring frames for unstable to stable warping. We leave it as the future work to seek a full-frame video stabilization solution. 

}
\section{Conclusion}\label{sec:conclusion}

We propose a novel method DUT for video stabilization by decomposing video stabilization into estimating and smoothing trajectories with DNNs for the first time. DUT learns to do video stabilization step by step in an unsupervised manner. Both quantitative and qualitative results on the popular NUS dataset confirms the performance of the proposed DUT methods. Moreover, it is light-weight and computationally efficient. More effort can be made to improve the performance of DUT further, \eg, devising an adaptive multi-plane segmentation network. We hope such a work can provide a new perceptive on unsupervised video stabilization learning, regards the difficulty in collecting paired stable-unstable videos.

\newpage

\bibliographystyle{IEEEtran}
\bibliography{egbib}

\end{document}